\documentclass[conference]{IEEEtran}
\IEEEoverridecommandlockouts

\usepackage{microtype}
\usepackage{graphicx}
\usepackage{booktabs} 

\usepackage{hyperref}
\usepackage{enumitem}
\usepackage{comment}

\renewcommand{\baselinestretch}{1.0}

\usepackage{amsmath}
\usepackage{amssymb}
\usepackage{mathtools}
\usepackage{amsthm}
\usepackage{xspace}
\usepackage{subfigure}
\usepackage{graphicx}

\usepackage{url}

\usepackage{xcolor}

\newcommand\fref{Figure~\ref}
\newcommand\tref{Table~\ref}

\newcommand{\minjia}[1]{\textcolor{orange}{Minjia:\ #1}}

\newcommand{\reza}[1]{\textcolor{red}{Reza:\ #1}}

\newcommand{\name}{DeepSpeed Inference\xspace}
\newcommand{\OURS}{DeepSpeed Inference\xspace}
\newcommand{\xyz}{DeepSpeed\xspace}

\newcommand{\zinf}{ZeRO-Inference\xspace}
\newcommand{\gpuinf}{DeepSpeed Transformer\xspace}
\newcommand{\sbi}{SBI-GeMM\xspace}
\newcommand{\pcc}{PCC\xspace}

\newcommand{\revision}[1]{\textcolor{black}{#1}}

\newcommand\extrafootertext[1]{%
    \bgroup
    \renewcommand\thefootnote{\fnsymbol{footnote}}%
    \renewcommand\thempfootnote{\fnsymbol{mpfootnote}}%
    \footnotetext[0]{#1}%
    \egroup
}

\newcommand{\moe}{DeepSpeed-MoE\xspace}

\theoremstyle{plain}

\theoremstyle{definition}

\theoremstyle{remark}

\begin{document}

\title{\OURS: Enabling Efficient Inference of Transformer Models at Unprecedented Scale}

\author{\IEEEauthorblockN{Reza Yazdani Aminabadi, Samyam Rajbhandari, Minjia Zhang, Ammar Ahmad Awan, \\ Cheng Li, Du Li,  Elton Zheng, Jeff Rasley, Shaden Smith, Olatunji Ruwase, Yuxiong He}
\IEEEauthorblockA{Microsoft Corporation\\
\{yazdani.reza,samyamr,minjiaz,ammar.awan,chengli1,du.li,elton.zheng,jeff.rasley,shaden.smith,olruwase,yuxhe\}@microsoft.com}}



\maketitle

\begin{abstract}
    The past several years have witnessed the success of transformer-based models, and their scale and application scenarios continue to grow aggressively. 
The current landscape of transformer models is increasingly diverse: the model size varies drastically with the largest being of hundred-billion parameters; the model characteristics differ due to the sparsity introduced by the Mixture-of-Experts; 
the target application scenarios can be latency-critical or throughput-oriented; the deployment hardware could be single- or multi-GPU systems with different types of memory and storage, etc.
With such increasing diversity and the fast-evolving pace of transformer models, designing a highly performant and efficient inference system is extremely challenging.

In this paper, we present \OURS, a comprehensive system solution for transformer model inference to address the above-mentioned challenges. 
\OURS consists of (1) a multi-GPU inference solution to minimize latency while maximizing the throughput of both dense and sparse transformer models when they fit in aggregate GPU memory, and (2) a heterogeneous inference solution that leverages CPU and NVMe memory in addition to the GPU memory and compute to enable high inference throughput with large models which do not fit in aggregate GPU memory. 

\OURS reduces latency by up to $7.3\times$ over the state-of-the-art for latency oriented scenarios and increases throughput by over 1.5x for throughput oriented scenarios. Moreover, it enables trillion parameter scale inference under real-time latency constraints by leveraging hundreds of GPUs, an unprecedented scale for inference.  It can inference $25\times$ larger models than with GPU-only solutions, while delivering a high  throughput of 84 TFLOPS (over $50\%$ of A6000 peak).

\end{abstract}

\section{Introduction}
\label{sec:introduction}

The past several years have witnessed the success of transformer-based models; their scale and application scenarios continue to grow aggressively. 
The current landscape of transformer models is increasingly diverse: the model size varies drastically with the largest being over trillion parameters; the model characteristics differ due to the sparsity introduced by the Mixture-of-Experts technique; 
the target application scenarios can be latency-critical or throughput-oriented; the deployment hardware could be single- or multi-GPU systems with different types of memory and storage, etc.
With such increasing diversity and the fast-evolving pace of transformer models, designing a highly performant and efficient inference system is extremely challenging.

\textbf{Latency Challenges:} Using a transformer based model for online scenarios in production requires meeting stringent latency requirements, and thus the batch sizes used are generally small. For small batch sizes, inference latency of a model is lower bounded by the time it takes to load all the model parameters from memory to registers. Meeting the latency requirements of a transformer model inference therefore is equivalent to achieving adequate overall memory bandwidth. 

Maximizing effective memory bandwidth at small batch sizes requires reading memory at near peak memory bandwidth for fully-connected (or, linear) layers which contain the majority of the model weights, while also minimizing kernel launch and data movement overhead of other operators like layernorm and softmax.
The GeMM implementations and other kernels designed for training  primarily focus on maximizing compute utilization at very large batch sizes and are sub-optimal for latency-critical inference.

In addition, for large models, even the peak memory bandwidth of a single device may not be sufficient to meet inference latency constraints. It requires aggregate memory bandwidth across multiple devices, which needs optimal parallelism strategies for partitioning the model computation across devices that minimizes the communication overhead across devices. Such parallelism strategies must cater to the variation in transformer architecture and hardware characteristics.

With respect to transformer architectures, we view them in two broad categories --- dense or sparse Mixture-of-Experts (MoE) transformer models.
The optimal parallelism strategy depends on the model architecture. For example, tensor and pipeline parallelism work only for dense transformers, while expert parallelism only works for sparse transformers. Moreover, transformer-based MoE models contain both dense and sparse transformer components, requiring a combination of different parallelism techniques to maximize the effective memory bandwidth across devices. Finally, with respect to hardware characteristics, modern clusters have heterogeneous network topology (eg. intra-node NVLink/NVSwitch and inter-node InfiniBand) which requires further consideration when developing parallelism strategies.

\textbf{Throughput Challenges}: 
\revision{
In addition to meeting latency, production workloads also have throughput targets to meet cost budget. At small batch sizes, where the workload is still memory bandwidth bound, the latency of the workload does not increase as long as the computation is entirely overlapped with model weight reads. Therefore, maximizing throughput while meeting the latency SLA requires not only maximizing the memory bandwidth utilization, but also overlapping compute with the model weight reads, and at the same time achieving high compute efficiency at small batch sizes to maximize the batch size whose compute can be overlapped with reading the model weights. Inference kernels must therefore achieve high memory bandwidth utilization and high compute utilization at small batch sizes, whereas training kernels simply need to achieve high compute utilization at much larger batch sizes. This makes developing inference kernels quite challenging.
}

Moreover, even for throughput bound scenarios with large batch sizes, inference workloads can differ from training workloads in terms of data flow and computation dependencies, requiring novel solutions to achieve high throughput. For example, generative transformers have dependencies between each generated token and the next token, which does not exist during training. As a result, it incurs higher memory requirement during inference to keep track of previously generated states. For large models that may require pipeline parallelism to fit the model in memory, this dependency across generated tokens also requires new pipeline schedules to keep all devices busy compared to training scenarios.

\textbf{Feasibility Challenges under Limited Resources}: A model with tens of billions of parameters is simply too large to fit in the memory of a single GPU device, and at hundreds of billions of parameters, it is too large to even fit in the aggregate GPU memory of a single node. For example, inferencing MT-NLG~530B~\cite{smith2022using} requires about 1TB of GPU memory just to fit the model for inference, requiring over three DGX-2 nodes consisting over two dozen of NVIDIA A100 40GB GPUs. Most data scientists simply do not have access to such GPU resources needed for inference of these massive models.


In this paper, we present \OURS, a comprehensive solution for transformer model inference designed to address the above challenges. \OURS consists of two components:

\subsubsection{\gpuinf}

\gpuinf, is a GPU only solution, designed to minimize latency while maximizing throughput for both dense and sparse transformer models. It achieves state-of-art latency and throughput for transformer models of all sizes and supports running on a single GPU or scaling to hundreds of GPUs to inference multi-trillion parameter models. 

The \gpuinf solution is a three-layered system architecture consisting of i) single GPU transformer kernels optimized for memory bandwidth utilization at low batch sizes and high throughput at large batch sizes, ii) many-GPU dense transformer layer, for scaling dense transformer models across GPUs using tensor-slicing and inference-optimized pipeline parallelism, and iii) massive-GPU scale sparse transformer layer, designed to scale MoE transformer layers to hundreds of GPUs using a combination of parallelism techniques and communication optimization strategies, while also minimizing single GPU sparse computation overhead using  optimized sparse kernels.

By taking this layered approach, where each layer addresses a unique aspect of the latency challenge: batch size, scaling dense models, and scaling sparse models, but are compatible and built on top of each other, we create a comprehensive system capable of achieving state-of-art latency and throughput at unprecedented scales for both dense and sparse transformer models despite the heterogeneity in batch size, model scale and model characteristics.

\subsubsection{\zinf}
\zinf is a heterogeneous GPU+CPU+NVMe based solution to address the memory challenge by enabling massive model inference with minimal GPU resources. In contrast to \gpuinf, for applications that are less latency sensitive but resource constrained, \zinf allows inference of models with hundreds of billions of parameters on a single or multiple GPUs as long as there is enough CPU or NVMe memory to store the model parameters. In addition, even when the model does fit in aggregate GPU memory, \zinf delivers better per GPU efficiency than \gpuinf by supporting much larger batch sizes. 

The main contributions of the paper are as follows:
\begin{itemize}
    \item Single GPU transformer kernels for minimizing latency and maximizing throughput via memory-bandwidth-centric fusion schedules and GeMM kernels (Sec.~\ref{sec:inference-kernels}). 
    \item A many-GPU dense transformer inference system that combines tensor-parallelism to minimize latency with inference optimized pipeline parallelism schedules and memory optimizations to maximize throughput (Sec. \ref{sec:many-gpu-dense}). 
    \item A massive-GPU sparse model inference system that combines: i) expert, data, and tensor parallelism, ii) novel communication optimizations and iii) sparse kernel optimizations to scale sparse inference on trillions of parameters across hundreds of GPUs (Sec. \ref{sec:optimizing_moe_inference_latency}). 
    \item \zinf that leverages CPU, NVMe and GPU memory along with GPU compute to make massive model inference accessible with limited resources (Sec.~\ref{subsec:zero_inf_section}). 
    \item Extensive evaluation of \OURS on a wide range of transformer models covering four aspects: i) For latency sensitive scenarios, \gpuinf shows latency reduction over state-of-the-art of up to $1.9\times$ for dense models (up to 175B parameters) and $7.2\times$ for sparse models (a 1T model under 25 ms), while scaling to 256 GPUs at 33\% peak memory bandwidth utilization, an unprecedented scale for inference. ii) For throughput oriented scenarios, \gpuinf demonstrates over $1.5\times$ gain over state-of-the-art (Sec.~\ref{s:throughput-oriented}). iii) Evaluation of \zinf on GPU resource constrained systems that shows \zinf can support inference with $25\times$ larger models than with GPU only solution while achieving over 50\% of peak hardware performance. (Sec.~\ref{subsec:zero_inf_eval}). iv) Performance analysis and breakdown of the different optimizations discussed throughout the paper (Sec.~\ref{s:perf-breakdown}). 
\end{itemize}

{\it Despite the diversity in transformer inference landscape, \OURS offers a versatile solution capable of achieving state-of-art latency and throughput for all variations of transformer model inference: dense or sparse, small or large batches, billions to  trillions of parameters, single GPU or across hundreds of GPUs. Furthermore, it democratizes access to large transformer inference by enabling them on systems with limited GPU resources.} \OURS is available for everyone to leverage though our open-source repository: \url{https://github.com/microsoft/DeepSpeed}. 

\section{Background and Related Work}
\label{sec:related}

\textit{a) Dense Transformer Models:} 
The size of transformer-based language models has been increasing by $10\times$ each year for the past few years, from models with a few hundred millions of parameters~\cite{devlin2018bert,liu2019roberta,radford2018improving,radford2019language}, to models with dozens of billions parameters~\cite{gpt-neox-20b,turing-nlp-17b}. Recently, GPT-3~175B~\cite{brown2020language}, Gopher~280B~\cite{rae2021scaling}, and MT-NLG~530B~\cite{smith2022using} further push this limit to hundreds of billions of parameters.
As larger models have demonstrated outstanding accuracy performance on various natural language understanding and generation tasks, this exponential growth in model scale would continue as long as the system and hardware technology could keep up with it.

\textit{b) Sparse Transformer Models:} The success of scaling dense language models has motivated researchers and practitioners to further propose the Mixture-of-Experts (MoE) technique which introduces sparsity in transformer models~\cite{lepikhin2020gshard}. Typical transformer model~\cite{vaswani2017attention} architectures have transformer blocks that consist of two consecutive sub-layers, a self-attention sub-layer followed by a position-wise feed-forward (FF) block. MoE models add conditional computation by replacing the feed-forward blocks with a Position-wise MoE layer with a variable number of experts and a top-k gating function. Increasing the number of experts allows scaling the MoE model size with only sublinear increase in computation cost, greatly reducing the training cost of the model. However, MoE models can be up to  $8\times$ larger than their quality-equivalent dense models \cite{switch-transformer,google_glam_blog,m6-t,deepspeed-moe-paper}, requiring much higher aggregate memory bandwidth to achieve comparable latency during inference.

\textit{c) System Technology for Memory and Performance Scaling:} 
The major challenge in scaling model sizes resides in the memory bottleneck. To satisfy the memory requirement, prior works have proposed various parallelism strategies to use the aggregated GPU memory within and across nodes.

{\it Tensor parallelism}~\cite{shoeybi2019megatron} splits model layers horizontally across GPU devices. 
As the number of GPU increases for tensor-slicing, two primary trade-offs show up: (i) lower compute granularity due to the smaller local problem size, and (ii) all-reduce communications in each transformer layer to aggregate the partial activations.
When scaling across node boundaries, the inter-node bandwidth is limited comparing to the fast intra-node connections, thus tensor parallelism can cause a significant latency degradation. 
In practice, tensor parallelism is often restricted to groups of GPUs sharing the high-bandwidth interconnect within a node (e.g., NVIDIA NVLink).

{\it Pipeline parallelism}~\cite{GPipe,harlap2018pipedream,megatron-2021} splits a model vertically into pipeline stages and use micro-batching to hide pipeline bubbles.
It only requires communication for data aggregation between adjacent pipeline stages, thus more efficient to scale across nodes.
However, model splitting and
micro-batching could pose functionality, performance and convergence related restrictions for pipeline parallelism. 

{\it ZeRO}~\cite{rajbhandari2020zero} takes a different approach and removes the memory redundancies in conventional data parallelism by partitioning model states across the data-parallel processes instead of replicating them.
3D parallelism ~\cite{deepspeed3dblog} combines data, tensor, and pipeline parallelism efficiently to scale to models of trillions of parameters.

\textit{Expert parallelism}~\cite{fedus2021switch} places different experts on different GPUs and executes them in parallel. Each expert only processes a subset of tokens on each expert based on a learned top-k gating function. The classic all-to-all communication primitive has been used to implement expert parallelism~\cite{rajbhandari2022deepspeed-moe, deepspeed-moe-paper, fedus2021switch}. 

The above parallelism strategies are mainly designed for maximizing training throughput and their effectiveness can be limited during inference because of insufficient parallelism with small batch sizes in inference.
Our work leverages these techniques
and applies innovative optimizations to make them effective and performant in inference.


\textit{d) Optimized Transformer Kernels:}
There is also a suite of work focused on accelerating the performance of transformer kernels \cite{deepspeed-bert, ivanov2021data, fang2021turbotransformers}.
\revision{
A record training time for BERT was accomplished with stochastic transformer kernels that fused operators and reduced activation memory to support large batch sizes~\cite{deepspeed-bert}.
Ianov~et~al.~\cite{ivanov2021data} use transformer dataflow graphs to fuse elementwise and reduction operators and accelerate training.
TurboTransformers~\cite{fang2021turbotransformers} similarly fuses elementwise and reduction operators
for transformer \textit{inference}. E.T.~\cite{et-transformer} combines fusion, custom GeMM, and pruning together to accelerate inference speed of Transformers.
The kernel optimizations presented in this work fuse a wider variety of operators,
such as head-wise transformation that requires additional data layout transformation and layers beyond the self-attention sublayers, such as the intermediate layers and MoE specific layers. In addition, the kernels presented in this work also support auto-regressive generative models that require KV-caching \cite{brown2020language} to be performant during inference, where as the above mentioned work do not consider support for KV-caching.}

\textit{e) DNN Inference Optimizations:} There has also been extensive work on optimizing DNN inference through platforms, libraries, compilation, and compression strategies. Several compilers and runtimes exist  to facilitate the deployment of models, such as TVM~\cite{chen2018tvm}, ONNXRuntime~\cite{ONNX_Runtime_developers_ONNX_Runtime_2018} and TensorRT~\cite{tensor-rt}. These platforms have been mostly focused on optimizing DNN models that can fit in a single GPU, such as small transformers with a few hundreds millions of parameters. 
In contrast, our work targets billion-scale or even trillion-scale transformers that do not easily fit on a single GPU device. The most related work to ours is FastTransformer~\cite{fasttransformer-nvidia}, which supports multi-GPU inference for transformer models, which we will provide a more detailed comparison in Section~\ref{sec:evaluation}. 
Finally, there has been numerous works that improves the deployment of DNN models through model compression techniques, such as distillation, quantization, and sparsification, which could reduce the computation time and memory consumption with a small accuracy trade-off. Our work is complimentary to these model compression techniques and can be combined together to boost performance further.

\begin{figure*}[ht!]
    \centering
    \vskip -1.5em
    \setlength{\abovecaptionskip}{-3pt}
    \resizebox{\textwidth}{!}{
    \subfigure[]{\includegraphics[width=0.35\textwidth]{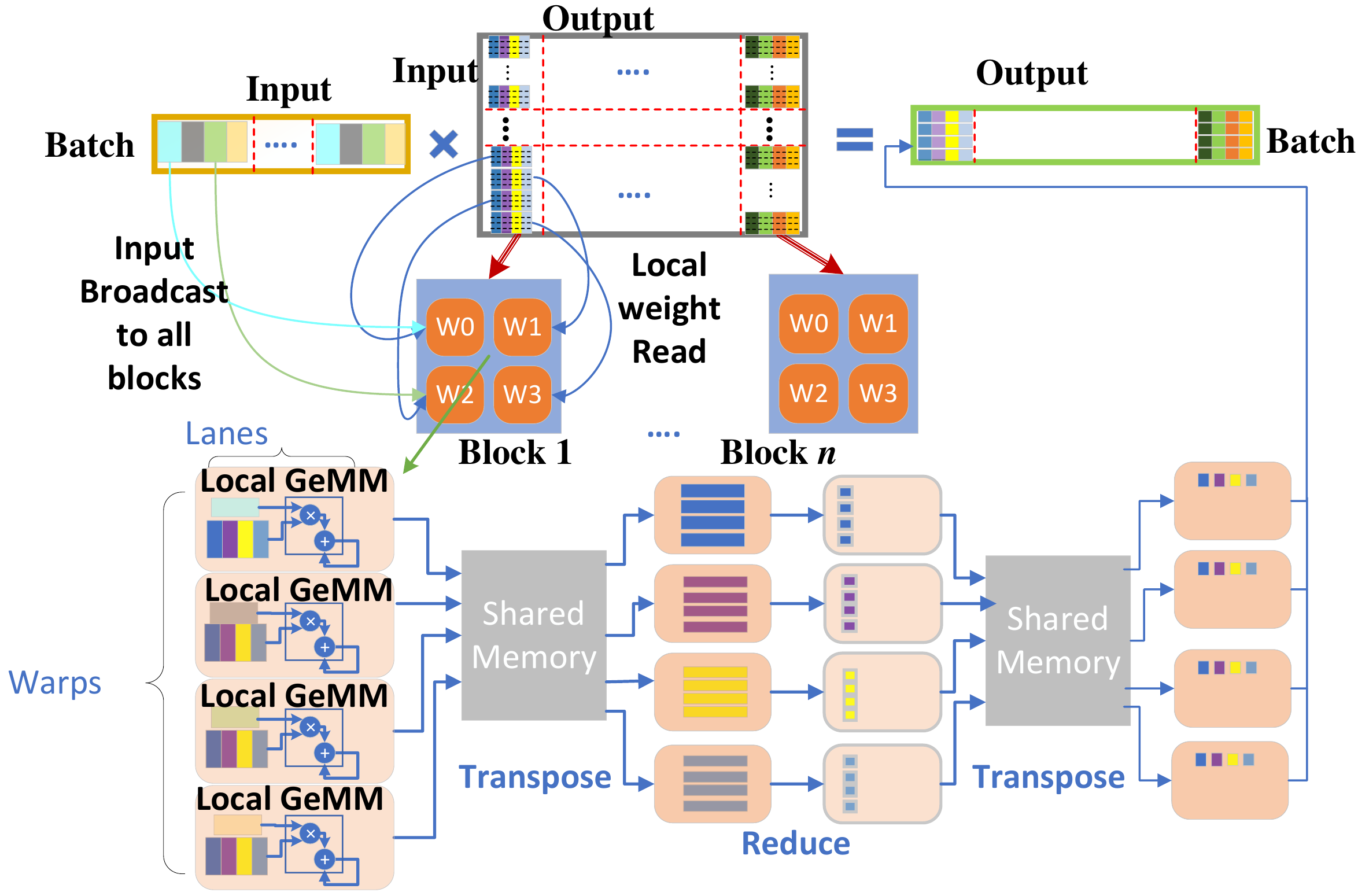}} 
    \subfigure[]{\includegraphics[width=0.30\textwidth]{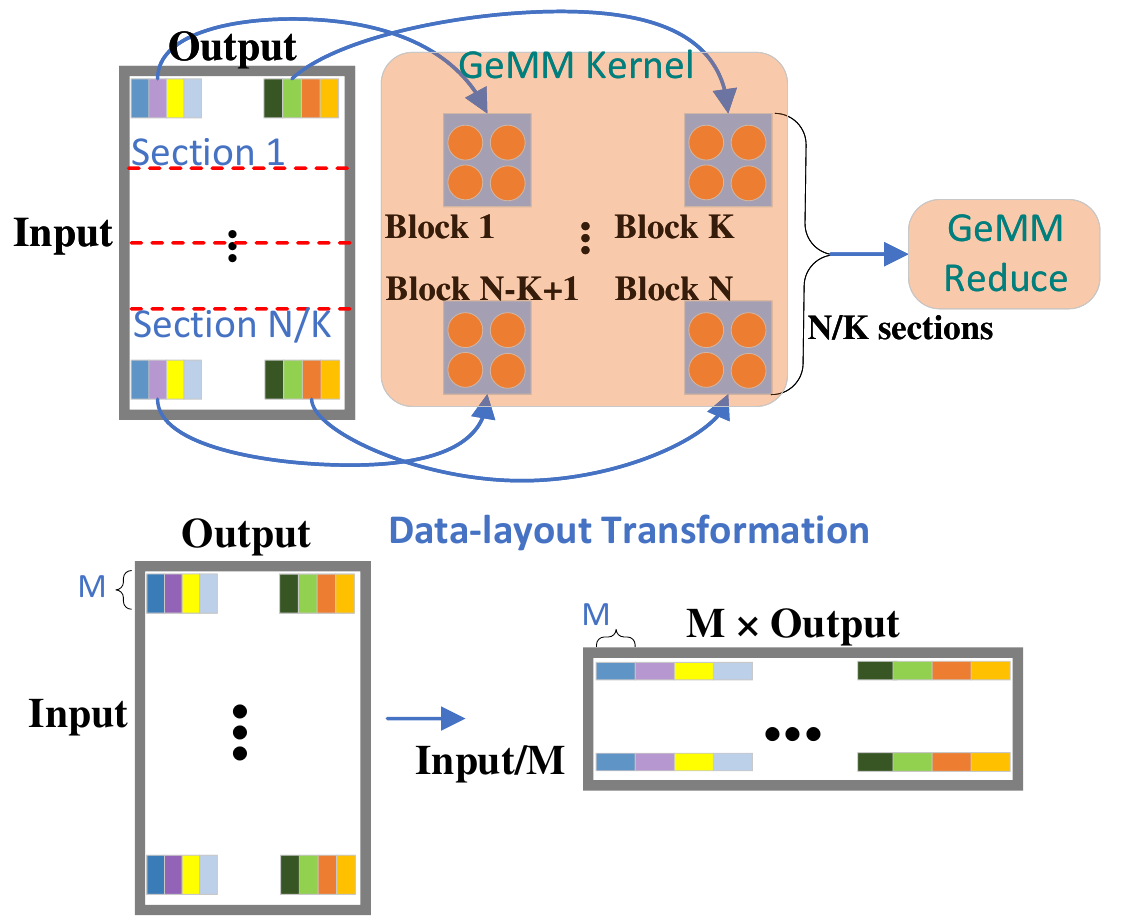}} 
    \subfigure[]{\includegraphics[width=0.35\textwidth]{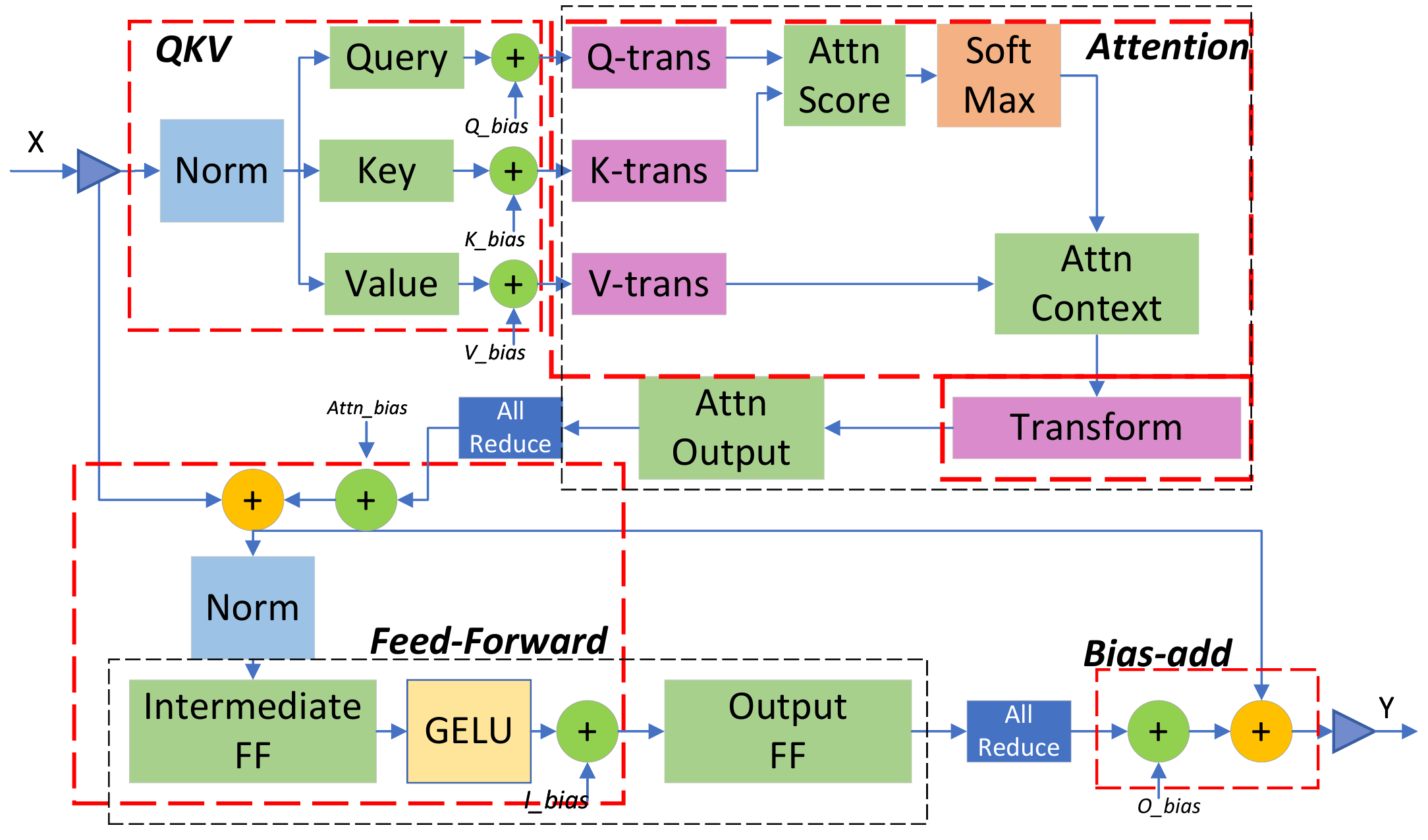}} 
    }
    \caption{ 
    (a) GeMM scheduling at different GPU architectural levels: threads-blocks, Warps and CUDA threads. Warps show different cooperative threads (32 threads), Lanes show the thread index at each Warp. 
    (b) GeMM modification to support 2-dimensional partitioning of the weight matrix and new data-layout. (c) Deep-Fusion strategy for the small-batch inference. }
    \label{fig:gemm-schedule}
    \vspace{-10pt}
\end{figure*}

\section{Inference-Optimized Transformer Kernels}\label{sec:inference-kernels}

In this part, we discuss the challenges, design, and optimizations for transformer kernels capable of achieving high-performance inference for both small and large batch sizes. 

\subsection{Inference Challenges on Different Batch Sizes}

As discussed in Sec.\ref{sec:introduction}, small batch performance is limited by the memory bandwidth utilization in reading model weights. 
There are three main challenges to optimizing for memory bandwidth at small-batch inference.
First, due to limited work at different kernels performing the operations of a transformer layer using small batch, 
inference performance suffers from the kernel-invocation overhead. 
Second, each kernel-invocation writes data to global memory which is read by GPU cores during the next kernel invocation, 
and this data-transfer between GPU cores and global memory adds an additional overhead. 
Finally, neither cuBLAS nor CUTLASS GeMM libraries are well tuned for extremely small batch sizes, and cannot achieve good memory-bandwidth utilization. 

Large-batch inference performance on the other-hand is limited by compute utilization, and while compute heavy operations like GeMM inside a transformer layer can achieve very good compute utilization using CUBLAS and CUTLASS libraries, the overall utilization can still be limited by the kernel launch overheads and data-transfers between GPU cores and global memory across different kernels other than GeMMs.

To address these challenges, we introduce two techniques: i) Deep-Fusion to reduce kernel-invocation and data-movement overheads by fusing multiple kernels beyond element-wise operations, and ii) A custom GeMM kernel designed for improving the memory bandwidth utilization when the batch size is relatively small while also allowing it to be fused using Deep-Fusion. We discuss these techniques in detail next.



\subsection{Deep-Fusion}
While operator fusion is a common technique used in deep learning to reduce kernel launch and data-movement overhead, it is limited primarily to element-wise operators  \cite{tensorflow_xla, chen2018tvm, ONNX_Runtime_developers_ONNX_Runtime_2018}. In contrast, transformer consists of operators like data layout transformations, reductions, and GeMMs which create data dependencies across thread blocks, making them difficult to fuse. This is because on GPU, if a data produced by a thread-block is consumed by a different one, a global memory synchronization is needed which invokes a new kernel. 

To avoid the need for a global synchronization, Deep-Fusion tiles the computation-space along dimensions of the iteration space which incur no cross-tile data-dependencies and executes them in parallel across different thread-blocks. The dimensions of the computation-space which does contain data dependencies are not tiled, and instead processed by the same thread-block. 

After this tiling, two operators can be fused using Deep-Fusion if each tile of the second operator depends on exactly one output tile of the first operator. By performing fusion at tile granularity, Deep-Fusion can fuse not only element-wise operations but also reductions, data transpositions, and GeMMs as long as there are no cross-tile dependencies. For example, all micro-operations in a layer-norm \cite{layernorm} can be tiled along the token dimension, while the reduction dimensions are processed within a tile. This allows all the micro-operations inside a layernorm to be fused into a single kernel despite consisting of multiple reduction operations. Furthermore, the data produced by each tile is either kept in registers or in shared memory when possible to allow for data-reuse across operators without incurring global memory data-transfer overheads.

%
\subsection{\sbi: Custom GeMM for Small Batch Size}

Our custom GeMM implementation is designed to be fusable with Deep-Fusion while achieving maximum memory bandwidth utilization. Its design can be viewed in three parts:  tiling strategies, cooperative-group reduction, and data-layout transformation for better memory bandwidth utilization.

\subsubsection{Tiling Strategies}
Fig.~\ref{fig:gemm-schedule}(a) depicts our GeMM scheduling for a skinny matrix multiplication. We first tile the computation along the output dimension. That allows us to implement GeMM using a single kernel by keeping the reduction within a tile. For small models, where the output dimension is too small to create enough parallel tiles to achieve good memory bandwidth, we tile the input dimension as well and implement GeMM as two kernels to allow for reduction across tiles.

\subsubsection{Cooperative-Group Reduction}
With the aforementioned tiling strategy, each warp in a thread block is responsible for producing a partially reduced result for a tile of outputs and a final reduction is needed across all the warps within the thread block. Usually this is implemented as a binary tree based reduction in shared memory which requires multiple warp-level synchronizations, thus creating a performance bottleneck. To avoid this, we perform a single data-layout transpose in shared memory such that partial results of the same output element are contiguous in memory, and can be reduced by a single warp using cooperative-group collectives directly in registers (See Fig.~\ref{fig:gemm-schedule}(a)). At the end, the first thread of each warp holds the final result and writes it to shared memory. The results in shared memory are contiguous, allowing for a coalesced write to global memory. 


\subsubsection{Leveraging Full Cache-line}
In GPU architecture, each L1 cache-line is 128 bytes, however a coalesced memory access with a single FP16 or INT8 element per thread in the warp cannot fully consume the full cache-line. Reading multiple elements per thread along the output dimension to address this issue reduces the number of parallel tiles which also hurts memory bandwidth. Therefore, our solution is to transpose the weight matrix during initialization such that $M$ rows for each column are contiguous in memory, allowing each thread to read $M$ elements along the input dimension (See Fig.~\ref{fig:gemm-schedule}(b)). We set $M$ as 2 for half precision and 4 for the INT8 data types considering a 128-byte cache line.

\subsection{Putting It Together}

{\it Small-batch Transformer Kernel}: Fig.~\ref{fig:gemm-schedule}.c shows the different components of a transformer layer, and the operations which are considered for Deep-Fusion 
in the small-batch inference case. As the figure shows, we fuse the operations inside a transformer layer at four main regions: 
1) the QKV GeMM and input layer-norm, 2) transposition plus attention, 3) post-attention layer-norm and intermediate GeMM, and 4) bias and residual addition. 
To support the fusion of GeMM with the rest of the operations in a single kernel for 3), 
we broadcast the input batch across the SMs and perform the same operations that come before GeMM, 
so that there is no need of communicating data between SMs for adding the GeMM schedule. 
We observe that in spite of replicating the work across SMs, we still gain performance benefit 
compared to the non-replicated, non-fused kernel implementation for the very small batch sizes.


{\it Large-batch Transfomer Kernel}: We follow the same fusion strategy as discussed above, 
with the difference that we use CUBLAS for GeMM operations, and keep them unfused. 

{\it Support for Different Data Types}: Our kernels support FP32, FP16 and INT8 data types for the GeMM operations. 
To support INT-8, we use CUTLASS~\cite{cutlass} INT8 GeMM implementation tuned for different batch sizes. We also add quantize operation before GeMM that we fuse using Deep-Fusion and de-quantization after GeMM that we fuse using CUTLASS's epilogue functionality.

{\it Eliminating Kernel Invocation Overhead via Cuda-Graph}:
\revision{For small to moderate sized models with small batch sizes, as we reduce the actual execution time of the kernels, the main latency bottleneck shifts from kernel execution to the kernel launch overhead on the CPU side.
To address this issue, we add the CUDA-Graph~\cite{cuda-graph} support in our inference pipeline. More specifically, we store the trace of the kernels the first time they are launched during the forward computation at inferencing and create the computation-graph that can be reused for the following requests, which largely eliminates the kernel launching overhead and substantially improves the performance.}

\section{Inference-Adapted Dense Transformer Models on Many-GPU Systems}
\label{sec:many-gpu-dense}

\begin{figure}[t]
\centering
\setlength{\abovecaptionskip}{0pt}
\subfigure[Baseline schedule with observed data dependencies.]{\includegraphics[width=0.85\columnwidth]{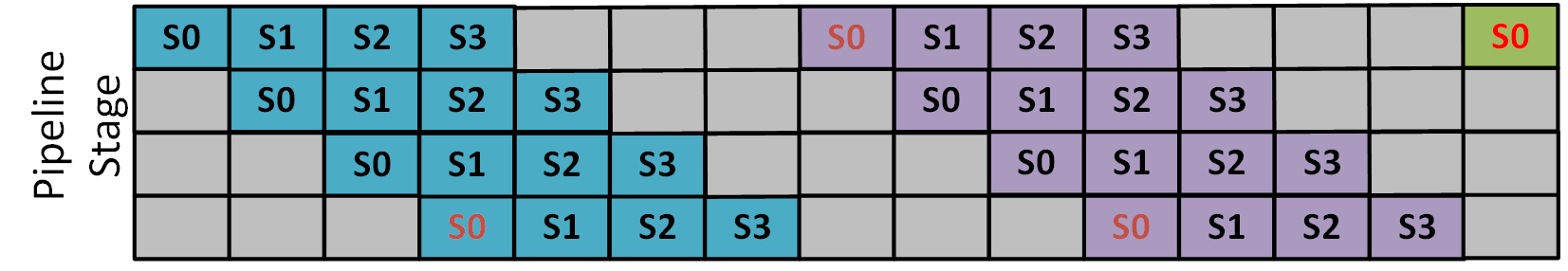}}
\subfigure[Pipeline schedule to hide data dependencies.]{\includegraphics[width=0.85\columnwidth]{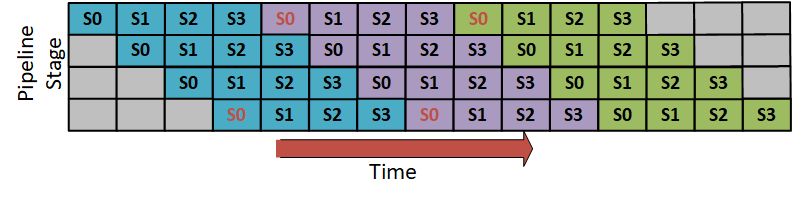}}

\caption{
A pipeline-parallel schedule for generating the first three tokens of four sequences $S0, \dots, S3$ using four pipeline stages. Sequence colors indicate the token being generated. Data dependencies exist between the first and last pipeline stages: we illustrate the dependencies for only \textcolor{red}{$S0$}. Gray blocks denote pipeline bubbles.
} 
\vskip -1.5em
\label{fig:pipe-schedule}
\end{figure}

\begin{figure}[t]
\centering
\includegraphics[width=1.0\columnwidth]{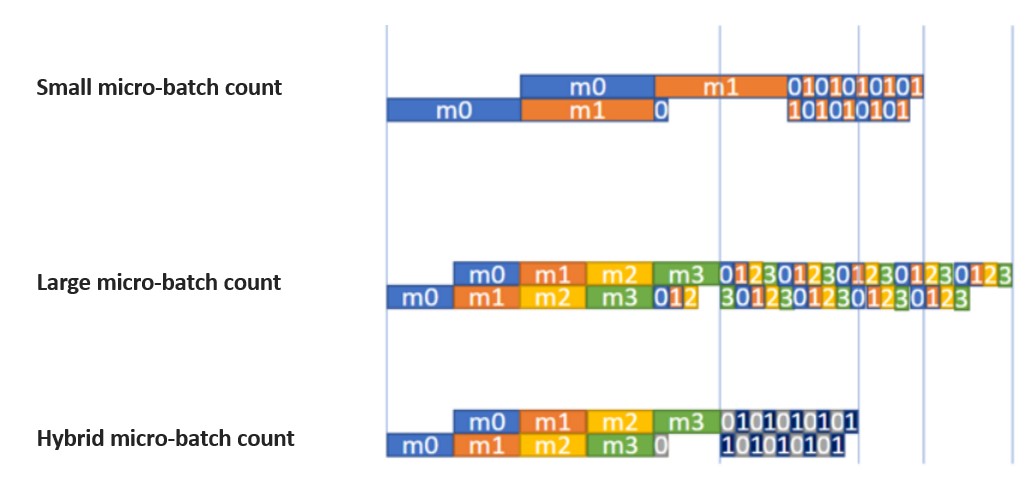}
\caption{Illustration of three different batch size combinations for prompt processing and token generation: A small micro-batch count reduces the latency of token-generation but prolongs the latency of prompt processing, and vice versa. By using a hybrid scheduling where different micro-batch counts are induced to different stages, the latency of both prompt processing and token-generation is reduced.}
\label{fig:hybrid_schedule}
    \vskip -1em
\end{figure}

This section presents the model parallelism techniques that we use on top of the single transformer kernels discussed in Sec. \ref{sec:introduction} with two goals: i) reducing latency further by leveraging aggregate memory bandwidth across GPUs and ii) increasing memory capacity by leveraging aggregate GPU memory across multiple nodes to fit massive models. While model parallelism is extensively studied in the context of training, there are unique challenges in inference, requiring new solutions.

\subsection{Aggregate Memory Bandwidth via Tensor Parallelism}
We leverage the aggregate memory bandwidth across multiple GPU devices via tensor-slicing parallelism (TP) from Megatron-LM~\cite{shoeybi2019megatron}. \name can automatically scale a dense transformer model to multiple devices by partitioning transformer operators across multiple devices while also adding appropriate communication operations needed across GPUs. Under the hood, it leverages the single GPU kernels to maximize per GPU memory bandwidth utilization, while using NCCL all-reduce collectives to perform the necessary across GPU communication as described in \cite{megatron-2021}. This allows \name to achieve excellent aggregate memory bandwidth utilization across several GPUs with a node. 
However, as discussed in Section~\ref{sec:related}, tensor slicing can not be scaled efficiently beyond a single node due to significant communication overhead. Thus to further scale to multi-node systems, \name uses pipeline parallelism.

\subsection{Aggregate Memory via Pipeline Parallelism --- Challenges}
\label{sec:pp-challenges}
As models exceed the memory capacity of a single node, we use pipeline parallelism (PP) \cite{GPipe, harlap2018pipedream}. 
Although PP does not help with the aggregate memory bandwidth since each micro-batch traverses the full depth of the model in sequence across the pipeline stages, it has smaller communication overhead (as discussed in Sec.~\ref{sec:related}) compared to TP, thus more efficient to scale across nodes.
However, applying PP in inference is non-trivial and requires different considerations from training:

    First, transformer decoders are autoregressive, i.e., the inputs to the model inference are previously-generated outputs. Therefore, when generating a sequence, the next token in the sequence is a function of the previous tokens.
    Existing training pipelines inference at the granularity of batches, and so batch boundaries are defined by the data dependencies of sequence generation. These data dependencies induce frequent pipeline bubbles that degrade inference performance (see Fig.~\ref{fig:pipe-schedule}).
    \revision{
    Second, autoregressive generation models have two distinct phases: i) prompt processing phase where the entire input prompt is processed to generate the first token and ii) token generation phase, where the results of the prompt processing is reused via KV-caching, and the new computation only depends on a single previously generated token. As the number of tokens processed in each phase is drastically different, they have different performance characteristics requiring different considerations.}
    
    Third, autoregressive inferencing caches the key and value activations of each transformer layer in order to avoid recomputation for each token. This activation memory scales with the number of sequences that are concurrently generated. In effect, inference performance for large transformer models can be limited by memory capacity. 

\subsection{Inference Optimized Pipeline Parallelism}
\label{pp_approach}
In order to overcome the inference-specific challenges, our approach includes three important aspects: scheduling, memory footprint reduction, and communication optimization.

\subsubsection{Hiding data dependencies and hybrid scheduling}
\label{pp_scheduling}
Suppose our goal is to inference a batch of $B$ sequences, $s_1, s_2, \dots, s_B$. We divide the sequence into groups of micro-batches, where one micro-batch is the unit of computation provided to each kernel invocation. A micro-batch progresses through the stages of the model pipeline until the next tokens are produced by the last stage of the model. If sequence $s_i$ does not terminate, the generated token will be used as the input for generating the next token.
Fig.~\ref{fig:pipe-schedule} illustrates our pipeline-parallel sequence generation schedule.
We set the number of micro-batches to the pipeline depth, $P$. Having at least $P$ micro-batches is critical to utilize all of the pipeline stages, but avoid additional micro-batches due to latency and memory costs of the larger batch size. However, we cannot repeatedly inference a batch of $P$ micro-batches without significant pipeline bubble overheads ruining efficiency. We avoid intermediate pipeline bubbles by dynamically queuing micro-batches of generated tokens until the sequences terminate. The resulting schedule amortizes the pipeline bubble over all generated tokens without allocating extra activations from a larger batch size.

\revision{
However, this is not sufficient to get the best performance. As mentioned in in Section~\ref{sec:pp-challenges}, autoregressive models, such as GPT-3, often consist of two stages that have different performance characteristics, and using the same micro-batch size for both stages is sub-optimal.}

\revision{
The prompt processing component of inference has a large number of tokens per sample that can saturate the GPU compute and the choice of micro-batches only affects the pipeline bubble but not the GPU execution time. However, for token generation, each sample has a single token, the total number of tokens across the entire micro-batch is small, and the kernel execution time is entirely memory bandwidth bound. That means, the execution time for a micro-batch does not change much with change in the size of micro-batch as most of the time is spent in fetching model parameters. However, the overall execution time is proportional to the number of micro-batches, as the forward pass on each micro-batch requires fetching the weights all over again. Therefore, efficient token generation requires minimizing the number of micro-batches while keeping it large enough to hide the pipeline bubble.}

\revision{To address the varying requirements for prompt processing and token generation, we adopt a hybrid scheduling strategy, where we use different number of micro-batches for the prompt processing and token-generation. Figure\ref{fig:hybrid_schedule} illustrates how the hybrid scheduling works. We use larger number of micro-batches during the prompt processing stage to minimize the pipeline bubble, while during the token generation phase, we reduce the number of micro-batches to reduces the overall execution time.}



\subsubsection{Offloading Activations to CPU Memory}
The cached key and value activation tensors have a predictable reuse pattern. The activations of sequence $s_i$ will not be used again until generating the next token of $s_i$.
When the allocated activation memory exceeds a threshold, we offload some activations from GPU to CPU memory while not in use. The saved GPU memory allows for larger batch sizes and enable better system utilization.

\subsubsection{Communication Optimization} Inference performance will ultimately degrade if the transformer kernels are stalled on communications for CPU memory offloading over the low-bandwidth PCIe. To avoid the stall we overlap the communication with computation, and more importantly we employ an architecture-aware communication optimized offloading strategy.  Most system architectures do not have a unique PCIe bus for each GPU and share a single link across two GPUs. To avoid contention between GPUs, odd-numbered GPUs offload activations for odd-numbered layers, while even-numbered GPUs offload activation for even-numbered layers. This is crucial to fully leverage the PCIe bandwidth. Scheduling odd and even layer offloading across GPUs prevents contention on the PCIe link, allowing each GPU to fully leverage the PCIe bandwidth when it needs to offload.

\begin{figure}[tbp]
\centering
\includegraphics[width=3.2in]{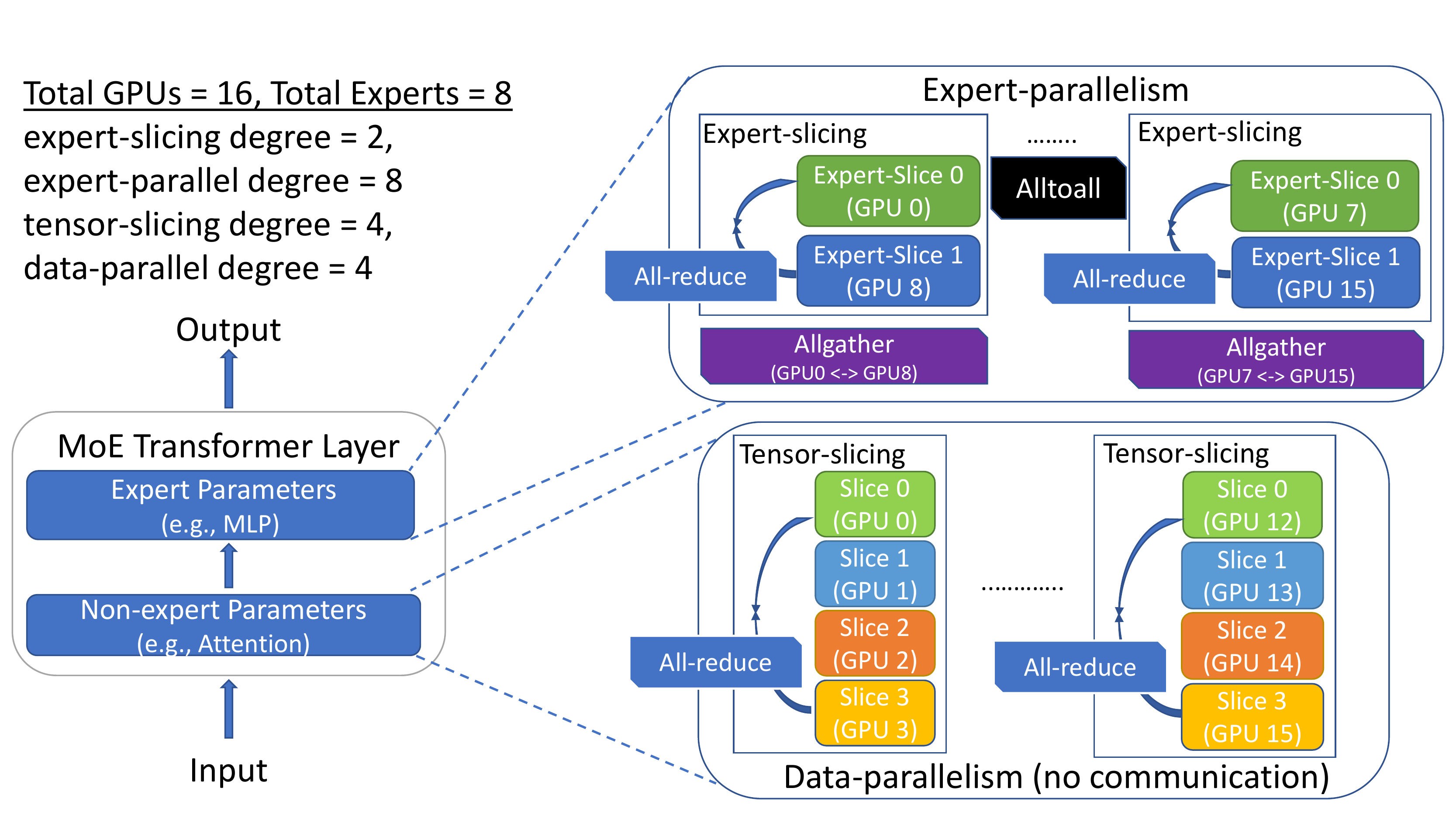}
\caption{Expert, data and tensor parallelism in \moe.}
\label{fig:inference-design}
    \vskip -1em
\end{figure}

\section{Massive Scale Sparse Model Inference}
\label{sec:optimizing_moe_inference_latency}

While the techniques developed so far enables \OURS to achieve state-of-art latency and throughput for dense transformer models, new considerations are necessary  for sparse transformer models that consist of both sparse and dense components. The key challenge is that on one hand, sparse models are much larger than quality equivalent dense models (Sec.~\ref{sec:related}), requiring much higher aggregate memory bandwidth to achieve latency comparable to quality equivalent dense models, and on the other hand it has a different computational structure than dense models, requiring different parallelism approaches compared to dense transformers \cite{rajbhandari2022deepspeed-moe}. 

In this section, we introduce a massive scale MoE-based transformer model inference system capable of addressing the above challenges. It is built on top of the dense components discussed before and consists of three main components:
\subsection{Orchestration of Tensor, Data, \&Expert Parallelism for MoE}

We use tensor parallelism, referred in Fig.~\ref{fig:inference-design} as tensor-slicing (for non-expert parameters) and expert-slicing (for expert parameters), to split individual parameters across multiple GPUs to leverage the aggregate memory bandwidth across GPUs. However, tensor parallelism can only scale efficiently to a few GPUs due to communication overhead and fine-grained parallelism. To address this, we use expert parallelism in conjunction with tensor parallelism to scale experts parameters to hundreds of GPUs. Expert parallelism does not reduce computation granularity of individual operators, therefore allowing our system to leverage aggregate memory bandwidth across hundreds of GPUs. To scale the non-expert computation to the same number of GPUs, we use data parallelism at no communication overhead.

\subsection{\pcc: Parallelism Coordinated Communication for MoE} \label{sec:opt-alltoall}

\begin{figure}[t]
\centering
\includegraphics[width=0.9\columnwidth]{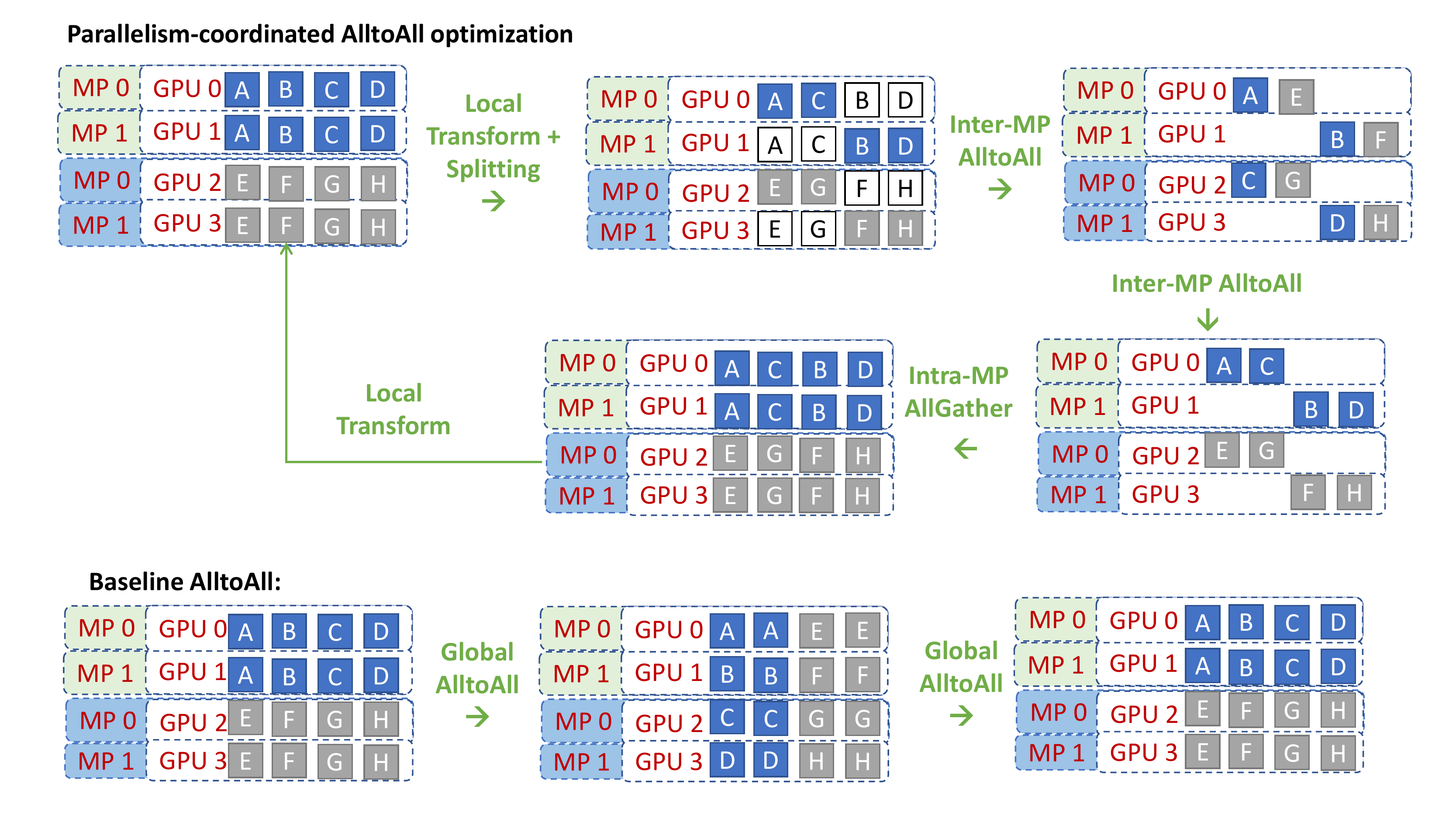}
\caption{The parallelism coordinated communication \revision{(PCC) optimization follows four steps: 1) local transformation and splitting of the original data, 2) Intra-tensor-model-parallel (MP) and inter-MP alltoall, followed by 3) intra-MP allgather, and 4) finally a local transform operation. Despite four steps, it is faster than the baseline alltoall shown in the bottom half of this illustration.}}
\label{fig:palltoall}
    \vskip -1em
\end{figure}

Expert parallelism places expert operators across GPUs and requires all-to-all communication between all expert-parallel GPUs. However, it is not efficient to scale expert parallelism to hundreds of devices needed for sparse model inference as the latency increases linearly with the increase in devices. Fortunately, when combining expert parallelism and tensor-slicing within a single model, there are opportunities for communication optimization that can reduce the communication latency. Note that tensor-slicing splits individual operators across GPUs and requires all-reduce between them. The all-reduce operation in tensor-slicing replicates data among the involved devices. When executing tensor-parallel operators followed by expert-parallel operators, this replication allows creating an optimized communication schedule for the all-to-all operator that does not require communicating between all the expert parallel processes: the all-to-all can happen within just the subset of devices that share the same tensor-slicing rank, since the data across tensor-parallel ranks are replicated (Fig.~\ref{fig:palltoall}). 
As a result, the latency of all-to-all is bounded by $O(p/L)$ instead of $O(p)$ where $L$ is the tensor-slicing parallelism degree and $p$ is the total number of GPU devices.

Similarly, when executing expert-parallel operators followed by tensor-slicing operators, the final all-to-all can be done in the same way, but this time followed by an allgather operator between tensor-parallel ranks to replicate the data needed by tensor-slicing (Fig.~\ref{fig:palltoall}). 
This reduces the latency overhead from $O(p)$ to $O(p/L) + O(L)$. 

This reduced latency overhead allows better scaling to a large number of devices. 
For example, when scaling to $128$ GPUs with $8$-way tensor-slicing and $128$-way expert parallelism, this approach reduces the latency overhead of the all-to-all from $(128C_1 + C_2)$ to $(16C_1+C_2)$ due to 8-way tensor-slicing, where $C_1$ and $C_2$ are some constants determined by point-to-point latency, message size, and bandwidth.

\subsection{Highly Optimized Computation Kernels for MoE}
\label{sec:opt-kernels}
\revision{
MoE-related computation consists of four major components: 
(1) a gating function that determines the assignment of tokens to experts, where the result is represented as a sparse tensor (a one-hot vector representing the assigned expert for each token in the sequence); (2) a sequence of sparse operators including a \textit{cumsum} operator to compute an inverse mapping from experts to token IDs (experts-to-token) using the previously mentioned token-to-expert one-hot vector; (3) a scatter operator to distribute tokens to its corresponding experts. This is implemented as a sparse einsum operator between the expert-to-token computed in the previous step and input tokens; and (4) a final sparse einsum based gather operation that re-distributes tokens processed at each expert back to their original ordering.
}

The sparse tensor representation in the gating function and sparse einsum operators introduce a significant latency overhead. The gating function includes numerous operations to create token-masks, select top-k experts, and perform cumulative-sum (\textit{cumsum}) to find the token-id going to each expert and sparse matrix-multiply, all of which are not only wasteful due to the sparse tenor representation, but also extremely slow due to many kernel call invocations. Moreover, the sparse einsums have a complexity of $S\times E \times M \times c_e$, where $S$ represents the total number of tokens, $E$ represents the number of experts, $M$ represents model hidden dimension, and $c_e$ represents expert capacity ($S$, $E$, and $M$ are the main complexity factors, while $c_e$ is normally very small). In this equation, $(E-1)$ out of $E$ operators for each token are multiplications and additions with zeros, since only one expert is typically selected to process $c_e$ tokens. This comes from the fact that generalizing the gating operations results in the einsums over several masking matrices or one-hot vectors that produce a lot of non-necessary computation with zeros to select the correct token for each expert. We optimize these operators using dense representation and kernel-fusion.

\revision{
We optimize each of the four steps in the gating function in the following way: 1) we replace the one-hot representation of the token to expert mapping using a table data-structure, greatly reducing the memory overhead from eliminating all the zeros in the one-hot vectors; 2) we create the inverse mapping (expert-to-tokens mapping table) from the tokens-to-expert mapping table by simply scanning though the token-to-expert table in parallel. 3) we replace the sparse einsum based scatter operation using a data-layout transformation that achieves the same result by first identifying the token IDs assigned to an expert using the expert-to-token mapping table created in the previous step, and then copying these tokens to the appropriate expert location; 4) after the tokens are processed by their corresponding experts, we use a similar data-layout transformation to replace the sparse einsum based gather operation.}

Using the data-layout transformation instead of sparse einsums reduces the complexity of these operations from $S\times E\times M\times c_e$ to $S\times M \times c_e$. \revision{ We use shared memory for data-layout transformations and fuse all but the final data-layout transformation together into a single kernel using basic fusion principles.} Combined, these optimizations result in over $6\times$ reduction in MoE kernel-related latency.

\section{Democratization of large model inference.}
\label{subsec:zero_inf_section}
\gpuinf needs the model to fit in aggregate GPU memory, requiring a large number of GPUs for large models. This is a barrier for many data scientists who lack access to large number of GPUs, e.g., dozens of GPUs are required to inference models like MT-NLG-530B.  To broaden access to large models, we propose \zinf which enables large model inference using as few as a single GPU. For non-latency sensitive applications, \zinf achieves high performance by leveraging DRAM and NVMe memories in addition to GPU memory and compute. Compared to a CPU only based solution, \zinf can achieve orders of magnitude higher throughput by efficiently exploiting the available GPU hardware. Moreover, it offers similar or even better throughput than \gpuinf by supporting larger batch sizes. We now discuss the design of \zinf and the performance optimizations that make it very efficient for throughput oriented inference.   

\subsection{\zinf Design}
\label{subsubsec:zero_inf_design}
\zinf utilizes available heterogeneous memory (i.e., GPU memory, DRAM, and NVMe) to satisfy the memory requirement of fitting massive models. This is motivated by the observation that environments with limited GPU resources are often equipped with terabytes of aggregate heterogeneous memory, which is sufficient to fit hundreds of billion-parameter models. \zinf builds on the offloading techniques of ZeRO-Infinity~\cite{rajbhandari2021zero},  and adapts them to inference.


\begin{table*}[t!]
\setlength{\abovecaptionskip}{0pt}
\centering{
    \begin{tabular}{|c|c|c|c|c|c|c|c|c|c|}
    \hline
    Name & \# params(B) & hidden dim (K) & \# layers  & \# attention heads 
    & Fig~\ref{f:single-gpu-inference} 
    & Fig~\ref{f:single-gpu-inference} 
    & Fig~\ref{fig:massive-model-latency_tput}
    & Fig~\ref{fig:zero_inf}\\ \hline
    GPT-[2, Neo, J, 13B] & 1.5, 2.7, 6, 13 & 1.6, 2.5, 4, 5 & 48, 32, 28, 40 & 25, 20, 32, 40 
    & TP=1 
    & N/A
    & N/A
    & N/A\\ \hline
    GPT-[NeoX, 50B, 87B] & 20, 50, 87 & 6, 8, 12, 12 & 44, 62, 48 & 64, 64, 96 
    &  N/A
    &  TP=2,4,8
    &  N/A
    & TP=1\\ \hline
    LM-175B & 175 & 12 & 96 & 96
    &  N/A
    &  TP=16
    &  TP=8, PP=2
    &  TP=1\\ \hline
    LM-530B & 530 & 20 & 105 & 128
    &  N/A
    &  N/A
    & TP=8,PP=5 
    & TP=1\\ \hline
\end{tabular}
}
\caption{Model configurations used for the dense model inference performance evaluation.}
\label{t:model-configs}
\end{table*}

\begin{table*}[htbp]
\footnotesize
\centering
\setlength{\abovecaptionskip}{0pt}
\begin{tabular}{rcccccccc}
\hline
 {Model} &  {Size (billions)}&  {\#Layers}&  {Hidden size}&  {MP degree} & {EP degree} & {Expert-slicing} & {\#GPUs}\\
\hline
 {1.3B+MoE-128} &  52& 24& 2048 & 1 & 128 & 1 & 128\\
 \hline
 {2.4B+MoE-128} &  107.7& 16& 3584 & 1 & 128 & 1 & 128\\
\hline
 {8B+MoE-128} &  349.0&  30&  4096  &  4&  128 & 1 & 128\\
\hline
 {24B+MoE-128} &  1064.9&  40&  8192&  8 &  128 & 2 & 256\\
\hline
 {47B+MoE-128} &  2024.0&  58&  8192&  8 &  128 & 2 & 256\\
\hline
\end{tabular}
\caption{Model configurations used for the sparse model inference performance evaluation. \revision{MP stands for model-parallelism. EP refers to expert-parallelism.}}
\label{tbl:moe-config}
\vskip -1em
\end{table*}

An important design decision is how to apportion GPU memory among model weights, inference inputs, and intermediate results. One approach is to pin as much of the model weights as possible into GPU memory, and fetch the remainder (from DRAM or NVMe) when needed for computation. A benefit of this approach is avoidance of the latency of fetching weights that are already pinned in GPU memory. However, this approach has two downsides: (i) it allows only small batch sizes which hurts efficiency, and (ii) the latency savings for hundred-billion parameter models are negligible since only a small fraction of the weights can fit in GPU memory anyway. 

\zinf adopts a different approach that pins the model weights either in DRAM (if large enough) or NVMe, and streams each layer into GPU memory for computation when needed. Despite the latency of fetching model weights over PCIe, \zinf is able to achieve high efficiency for two reasons. First, by limiting GPU memory usage of the model to one or a few layers of weights, \zinf is able to use large batch sizes for inference. Second, a large model layer requires significant amount of compute, especially given their long input sequence length (e.g., 2048). For example, one GPT3-175B layer requires about 7 TFlops to process an input of batch size 1. Therefore, large batch sizes cause compute time to dominate the latency of fetching model weights, which ultimately improves efficiency. In summary, \zinf's strategy to utilize GPU memory to support large batch sizes results in high performance inference for large models.   

\subsection{Performance Optimizations}
\label{subsubsec:zero_inf_efficiency}
 \zinf implements two optimizations to further mitigate the impact of fetching model weights from DRAM or NVMe for inference computations.
 
{\it Prefetching}: \zinf prefetches a configurable number of layers ahead of use, overlapping with computation of the current layer. Prefetching gives the flexibility to improve throughput at the cost of a configurable increase in GPU memory consumption.

{\it Multi-GPU PCI-e bandwith utilization}: In multi-GPU scenarios, the aggregate PCI-e bandwidth is used to reduce the layer transfer time by having each GPU only fetch a partition of the layer and then aggregating partitions over the much faster GPU-GPU interconnect.  

\revision{Beyond the above optimizations, \zinf also performs several other efficiency optimizations to achieve close to peak NVMe IO bandwidth, such as bulk read/write requests for asynchronous completion, aggressive parallelization of I/O requests, work scheduling, memory pinning, and avoiding data copy. However, we do not claim novelty on those optimizations as they were introduced by prior work~\cite{rajbhandari2021zero}.}

\section{Performance Evaluation}
\label{sec:evaluation}

We present an extensive evaluation of \OURS covering four aspects. i) For latency sensitive applications, \OURS achieves up to $1.9\times$ and $7.3\times$ lower latency than state-of-art for a wide range of dense models with hundreds of billions of parameters, and sparse models with trillions of parameters scaling to hundreds of GPUs. ii) For throughput-oriented inference of massive models, \OURS achieves up to $1.5\times$ higher throughput. iii) On resource constrained systems \OURS enables inference of $25\times$ larger models than GPU-only solution (530B vs 20B) while achieving over 50\% of peak hardware performance, democratizing large-model inference with limited GPU resources. iv) We present a performance breakdown to zoom into the contributions of individual optimizations. \vspace{-0.5em}

\subsection{Evaluation Methodology}
\label{subsec:eval-methodology}

\begin{figure*}[h!]
\centering
\setlength{\abovecaptionskip}{0pt}
\includegraphics[width=6in,height=2.2in]{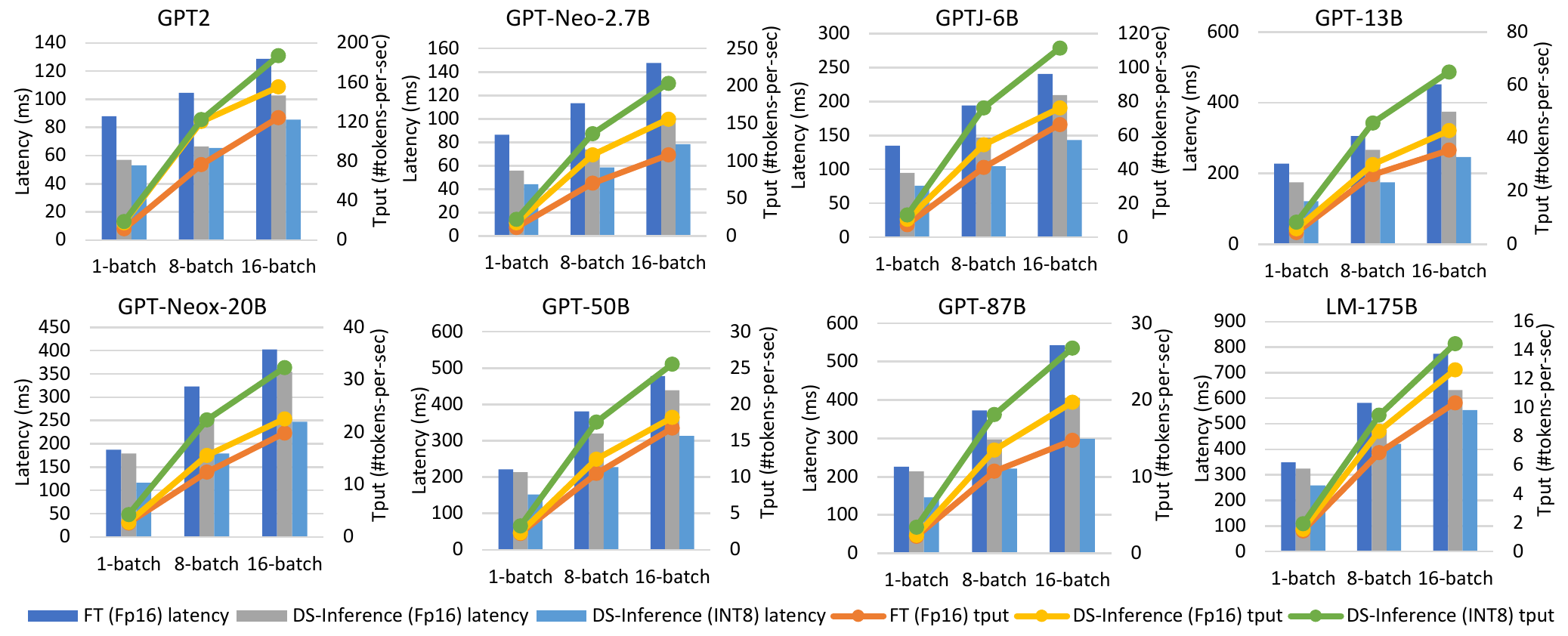}

\caption{Latency and throughput comparison of \gpuinf with FasterTransformer~\cite{fasttransformer-nvidia} for different models and batch sizes.}
\label{f:single-gpu-inference}
\end{figure*}

\subsubsection{Baseline} For dense models, we use FasterTransformer (FT)~\cite{fasttransformer-nvidia}, an efficient implementation of transformer models provided by NVIDIA. For experiments of sparse models, we use a full-featured distributed PyTorch implementation that supports both tensor and expert parallelism~\cite{kim2021scalable}. 

\subsubsection{Metrics} We use three performance metrics: (i) {\it latency}, i.e., end-to-end output generation time for a batch of input prompts, (ii) {\it token throughput}, i.e., tokens-per-second processed, and (iii) {\it compute throughput}, i.e., TFLOPS per GPU. 

\subsubsection{Workloads} For the performance evaluation, we focus on evaluating GPT-style transformer-based decoder models~\cite{brown2020language}, where we vary the hidden dimension, the number of transformer layers, and attention heads based on the GPT-3 paper as well as its publicly available variants to cover a wide range of model configurations and different number of parameters. Table~\ref{t:model-configs} elaborates the model architectures. For sparse MoE models, we further vary the expert degree to cover models ranging from 52B parameters to 2 trillion parameters. Sparse model configurations are shown in~\tref{tbl:moe-config}. Since generative text language models like GPT-3 produces tokens based on a prompt, which is the text given to the model to be completed, we measure the latency of generating 8 tokens with an input prompt of 128 tokens for dense models varying batch sizes, which reflects scenarios that correspond to more latency-sensitive applications. For the sparse MoE model, we measure the per-token latency by generating 100 tokens at a time with a prompt of 128 tokens and batch size 8. For throughput oriented applications, we measure the performance with an input prompt of 512 tokens while generating 50 tokens at a time. For resource constrained systems, we measure the compute throughput using maximum batch size possible for generating a single token.


\subsubsection{Testbeds} We conduct our experiments on:
a cluster of up to $256$ NVIDIA Ampere A100 40GB GPUs (32 $8\times$A100 DGX boxes~\cite{dgxa100}),
a lambda A6000 workstation~\cite{lambda} ($2\times$A6000-$48$GB-GPU, $256$GB DRAM, and $2$TB NVME) and a DGX2 V100 server~\cite{dgx2v100}  ($16\times$V100-$32$GB-SXM-GPU, $1500$GB DRAM, and $30$TB NVME). \vspace{-0.5em}
\subsection{Evaluation of \OURS for Latency Sensitive Workloads}
\label{s:latency-sensitive}


\OURS provides a comprehensive system solution to support fast inference of dense models over 530B parameters as well sparse models that have more than 2 trillion parameters at unprecedented scale. 
\begin{figure*}[htbp]
\centering
\includegraphics[width=\textwidth]{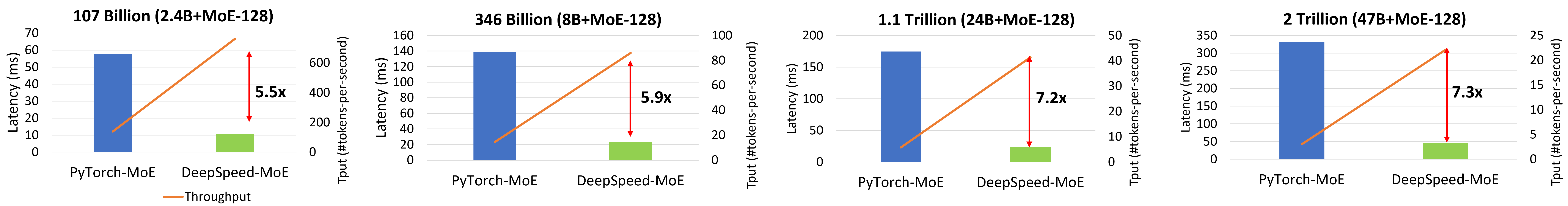}
\caption{Latency and throughput improvement offered by \moe over baseline on 256 GPUs. Throughput shown is per GPU and \revision{the speedup values along the arrows refer to improvement in latency.}} 

\label{fig:moe-vs-moe-new}
\vskip -1.5em
\end{figure*}

\subsubsection{Dense Model Evaluation} Fig.~\ref{f:single-gpu-inference} shows the latency and throughput improvements of \OURS on up to 175B parameter models running with up to 16-way tensor parallelism (see Tab.~\ref{t:model-configs}). In particular, we compare both FP16 (\xyz-FP16) and INT8 (\xyz-INT8) implementations of \OURS with the FasterTransformer FP16 baseline (FT-FP16)~\footnote{\revision{As the time of writing, FasterTransformer only supports INT8 computation for Transformer models with just the encoders, e.g., BERT, but not decoders used in state-of-the-art large-scale Transformer models such as GPT3~\cite{brown2020language}.}} Both the baseline and \OURS uses identical TP strategy so all the latency differences in these results come from the differences in kernel implementations described below.

{\it Small Batch Sizes} For small batch size, \xyz-FP16 achieves a speedup of up to $1.55\times$ over the baseline. The performance improvements for both single GPU and multi-GPU configs are primarily due to deep-fusion and custom GeMMs. The latency reduction is the largest for the smallest model sizes, as they have the largest kernel-launch overhead due to limited work per kernel, and worst GeMM memory bandwidth utilization from CUBLAS as they are not optimized for small and skinny GeMMs. \xyz-INT8 enables a further performance boost of up to $1.95\times$ over the FP16 baseline by reducing the overall size of the parameters in half compared to FP16. 

{\it Larger Batch Sizes} For larger batch sizes, \xyz-FP16 reduces the latency by up to $1.57\times$ over the baseline, and up to $1.93\times$ using \xyz-INT8. The primary source of performance improvement for \xyz-FP16 is the reduction of non-GeMM data-movement overhead via deep-fusion. As batch size increases, the GeMM becomes much more efficient, and the latency of the GeMM operators only increases sub-linearly with the batch size in this modest batch size regime. However,the latency of the non-GeMM operations increase linearly due to proportional increase in data movement from GPU memory, making it a bigger fraction of the overall latency. Deep-fusion reduces this data movement by keeping intermediate data for fused operators in shared memory or registers to achieve higher performance. The \xyz-INT8 further improves upon the \xyz-FP16 performance by utilizing the higher peak of the INT8 tensor-cores compared to FP16.

\subsubsection{Sparse Model Evaluation}
\label{subsec:moe-perf}
Fig.~\ref{fig:moe-vs-moe-new} shows the single output token generation latency and throughput of serving 100B to 2T MoE models with up to 256 GPUs with and without \moe.  
Compared to baseline, \moe achieves \revision{better performance than the state-of-the-art}, with up to $7.3\times$ reduction in latency. 
\revision{To have a fair comparison, the configuration for data/tensor/expert parallelism is the same for both the baseline and \OURS-MoE. The main differences are optimizations that \OURS has, such as expert-slicing, parallelism coordinated all-to-all and MoE-specific kernels, but the PyTorch-MoE baseline does not.}
By effectively exploiting hundreds of GPUs in parallel, \moe achieves an unprecedented scale for inference at incredibly low latency - a staggering trillion parameter MoE model can be served under 25ms by leveraging an aggregate GPU memory bandwidth of 128 TB/sec (33 \% of peak memory bandwidth), making it possible to serve such a massive model even in extremely interactive online applications. 

While we realize that 33\% compute utilization on 256 GPUs would be a fairly low for a compute bound application such as training with high arithmetic intensity, a 33\% memory bandwidth utilization for enabling a low latency massive model inference with virtually no arithmetic intensity is an unprecedented result due to the intensive communication required in such scenarios.

\begin{figure}[htbp]
\centering

\includegraphics[width=0.90\columnwidth]{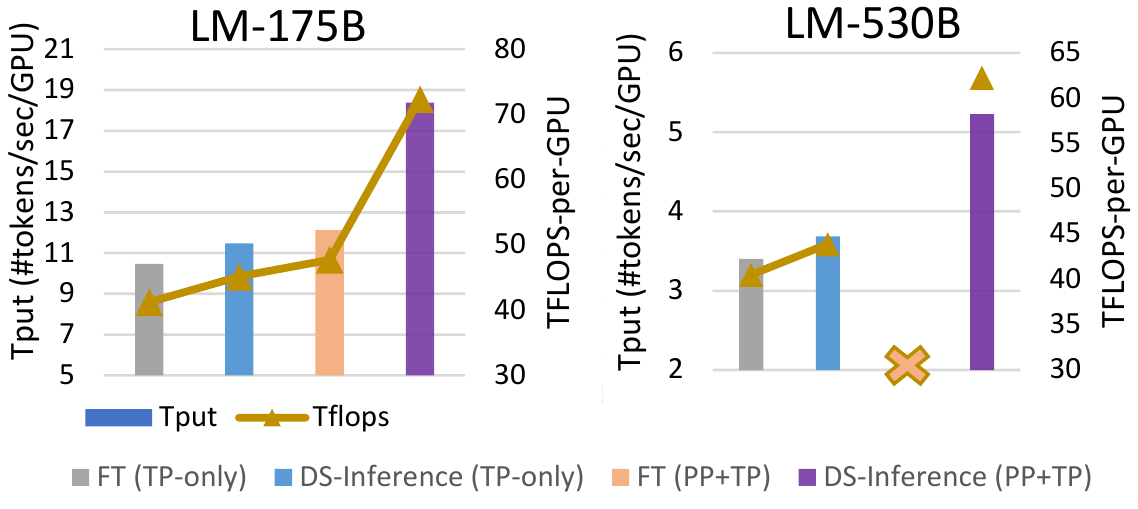}
\caption{ Throughput comparison of \gpuinf with FT for 175B and 530B models on 16 and 40 GPUs. We run with batch sizes that give the best performance for each configuration. }
\label{fig:massive-model-latency_tput}
\vskip -1.5em
\end{figure}

\begin{figure*}[ht]
    \centering
    \setlength{\abovecaptionskip}{-2pt}
    \resizebox{\textwidth}{!}{
    \subfigure[]{\includegraphics[width=0.25\textwidth]{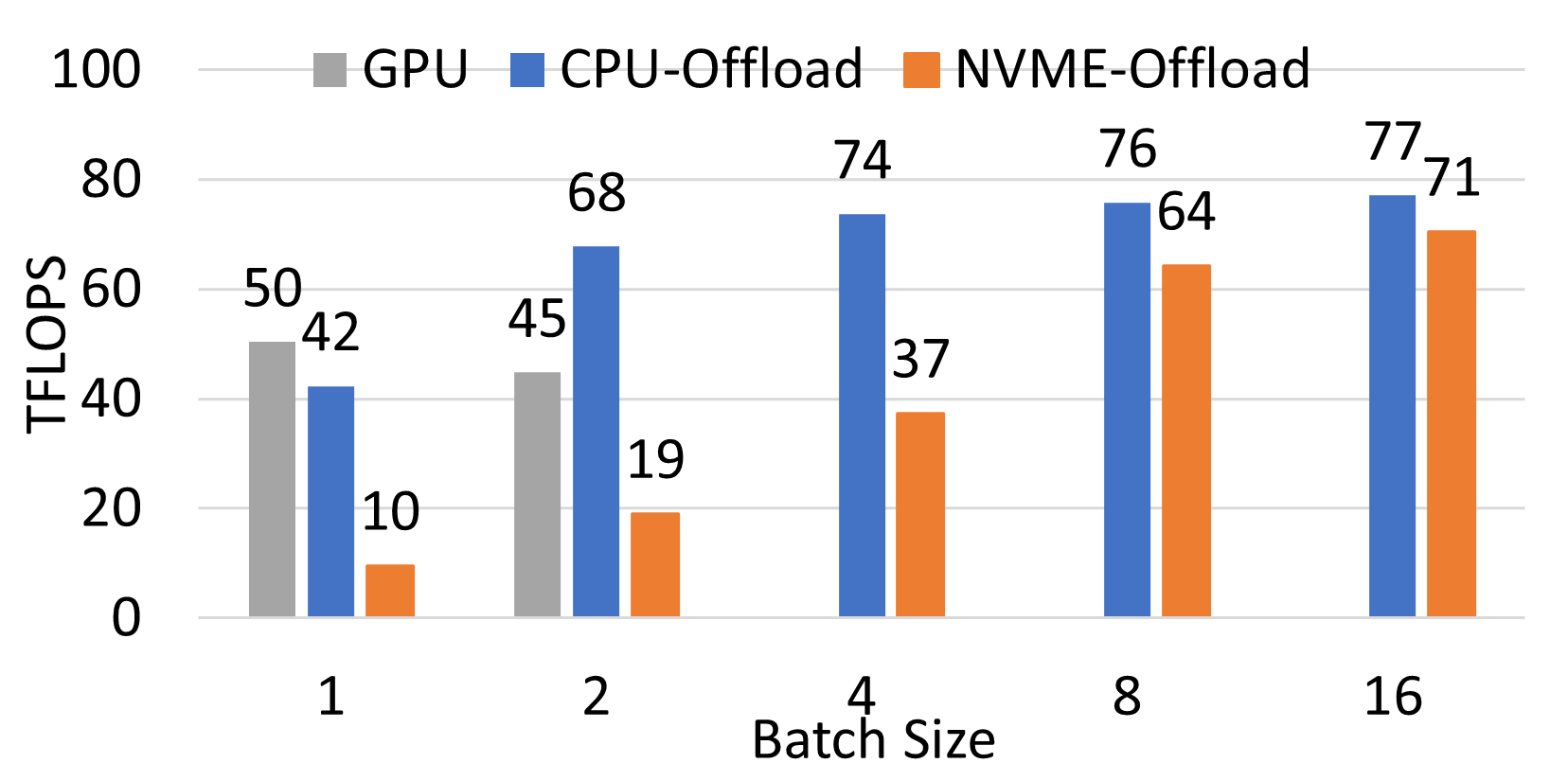}} 
    \subfigure[]{\includegraphics[width=0.25\textwidth]{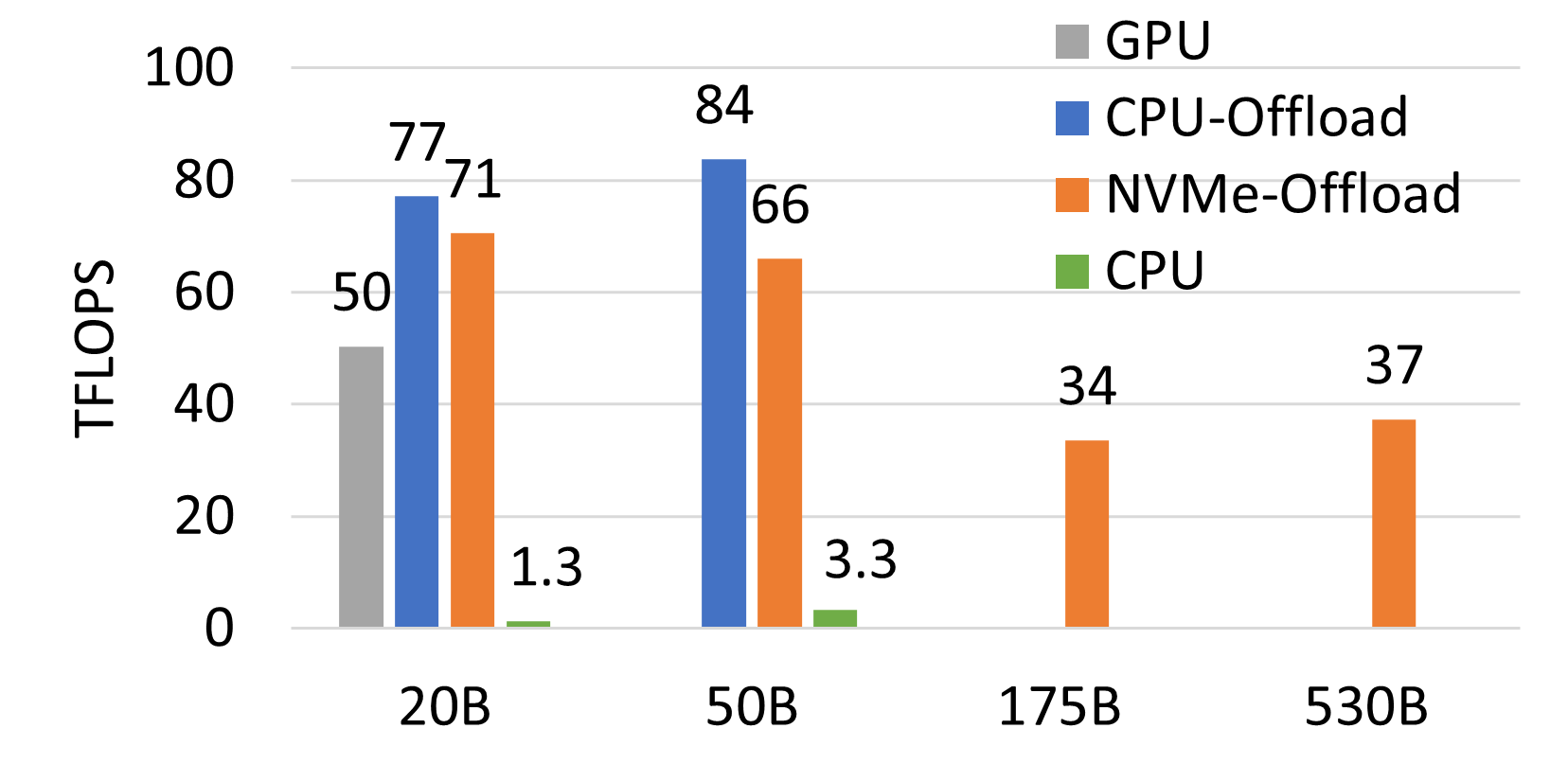}} 
    \subfigure[]{\includegraphics[width=0.25\textwidth]{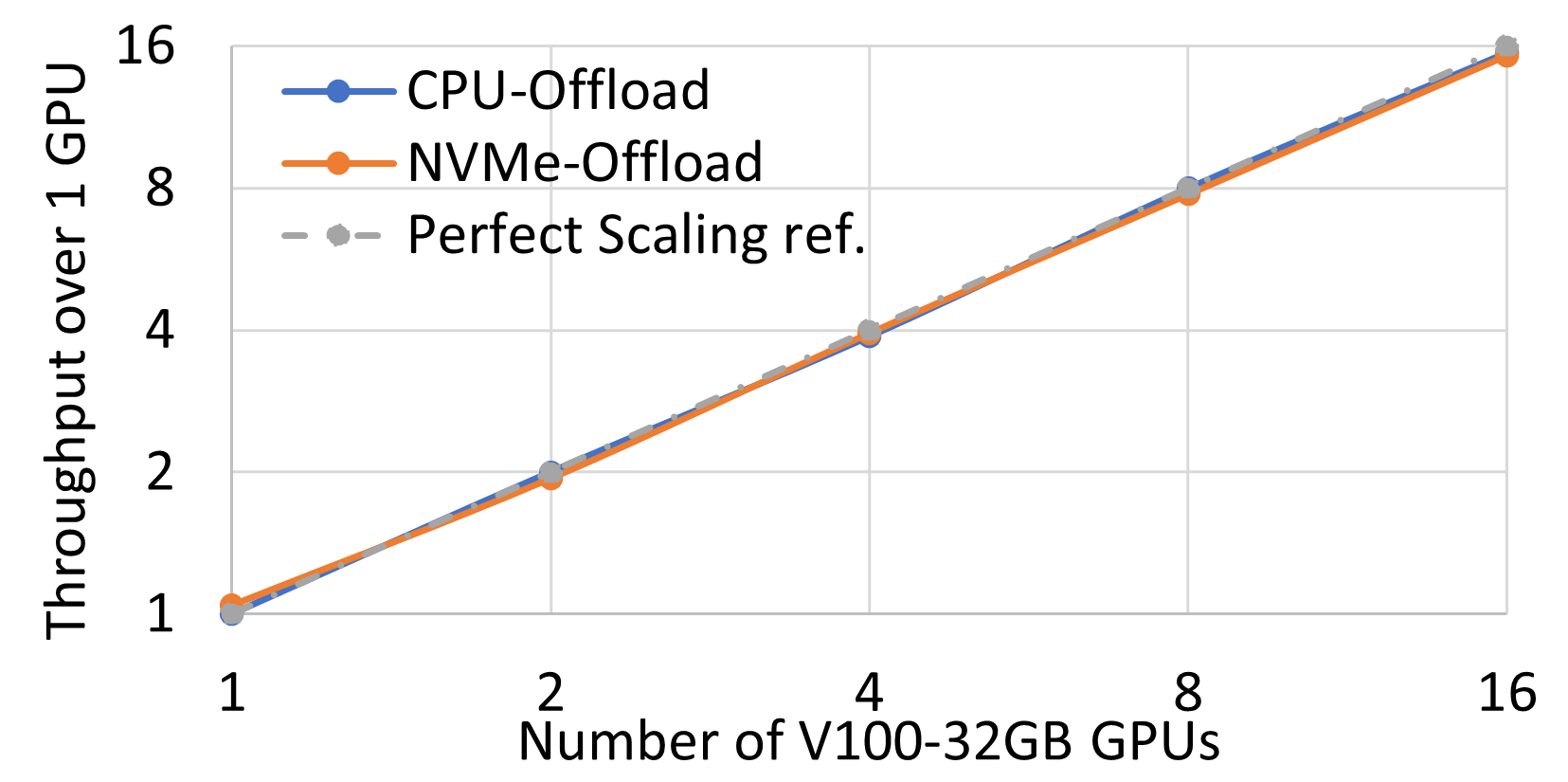}} 
    }
    \caption{ (a) Throughput of GPT-NeoX-20B across batch sizes on a A6000 GPU. (b) Throughput across models on a A6000 GPU. (c) Throughput of GPT-50B using up to $16$ GPUs over a single GPU (67 TFLOPS, 53\% of peak) on the DGX2 V100.}
    \label{fig:zero_inf}
    \vspace{-10pt}
\end{figure*}

%
\subsection{Throughput Oriented Massive Model Inference}
\label{s:throughput-oriented}
Massive models are capable of processing large input prompts and generating large number of coherent tokens.
In some applications (e.g., offline query rewriting in web-scale search and recommendation systems), this token generation process can be less latency focused and more throughput oriented. In this sub-section we show throughput improvement of \OURS for massive model inference.

Fig.\ref{fig:massive-model-latency_tput} shows that \OURS achieves $1.51\times$ throughput improvement over the best FasterTransformer (FT) configuration for the GPT-3 175B model running on two nodes ($2\times8$ A100). This improvement comes from our improved pipeline parallelism schedule, and ability to run much larger batch sizes using memory optimization and communication minimization strategies described in Sec.~\ref{sec:many-gpu-dense}. For the 530B, we could not run FT using a combination of TP and PP without crashing, but compared to the TP only version of FT, \OURS achieves over $1.53\times$ throughput improvement running on 5 nodes.

\vspace{-0.5em}
\begin{figure*}[ht]
    \centering
    \vskip -0.5em
    \setlength{\abovecaptionskip}{-2pt}
    \resizebox{\textwidth}{!}{
    \subfigure[]{\includegraphics[width=0.25\textwidth]{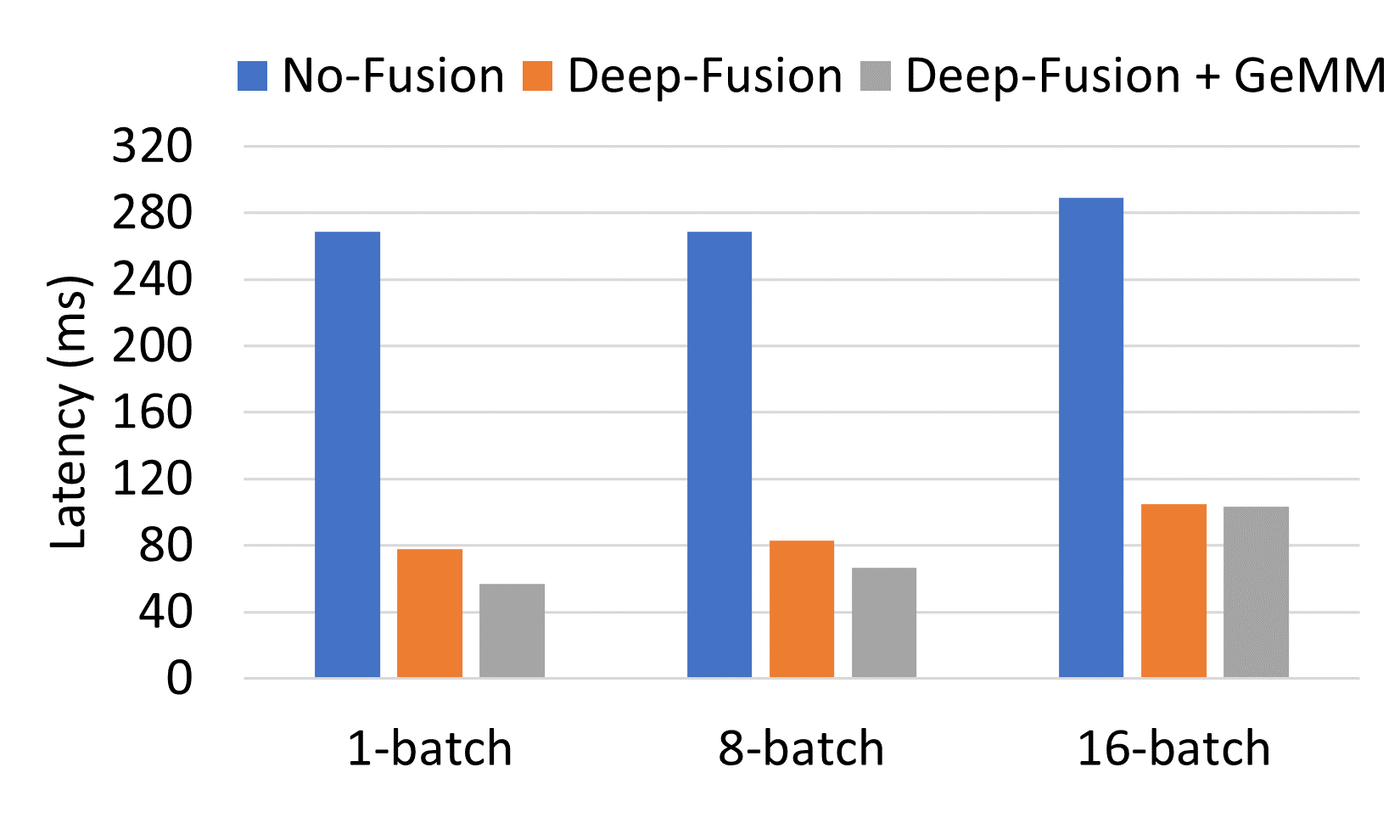}} 
    \subfigure[]{\includegraphics[width=0.25\textwidth]{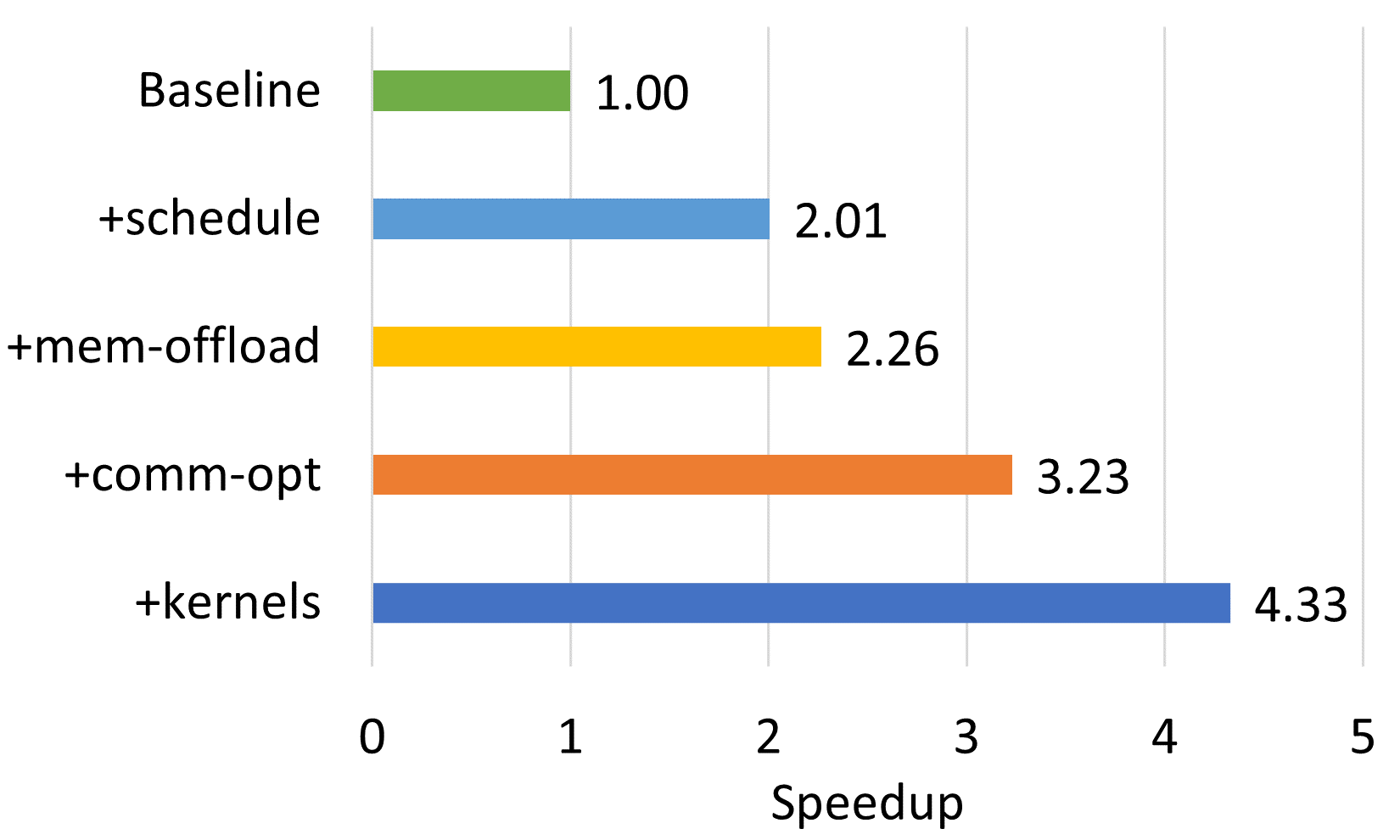}} 
    \subfigure[]{\includegraphics[width=0.25\textwidth]{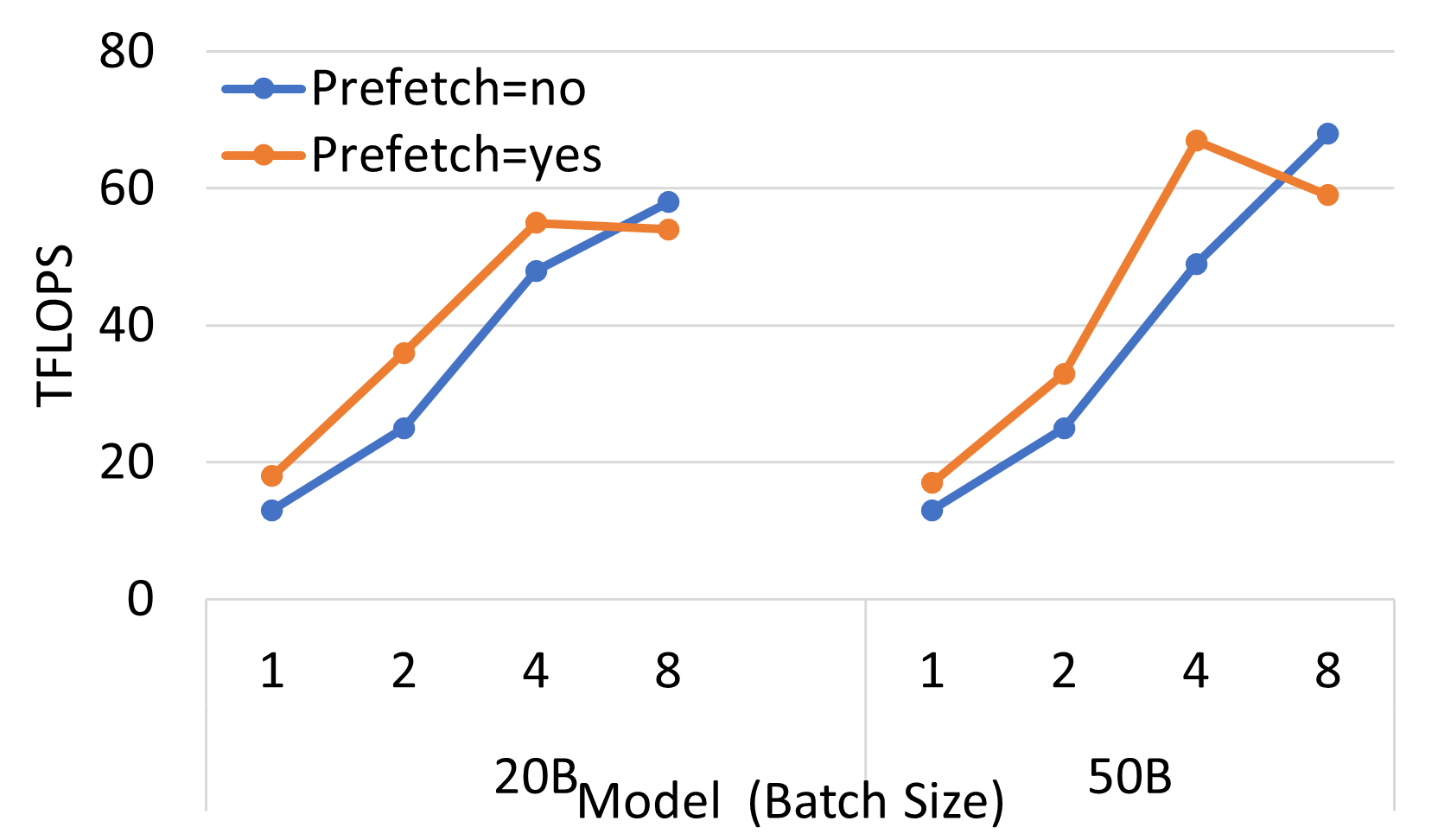}} 
    }
    \caption{ (a) Benefit of the Deep-Fusion and optimized GeMM over Megatron baseline for the GPT2 model. (b) Throughput improvement with different pipeline parallelism optimizations for 530B Model. (c) Impact of  prefetching on \zinf performance on a single V100 GPU.}
    \label{fig:perf-analysis}
    \vspace{-10pt}
\end{figure*}

\begin{figure}[t]
\centering
\includegraphics[width=0.8\columnwidth]{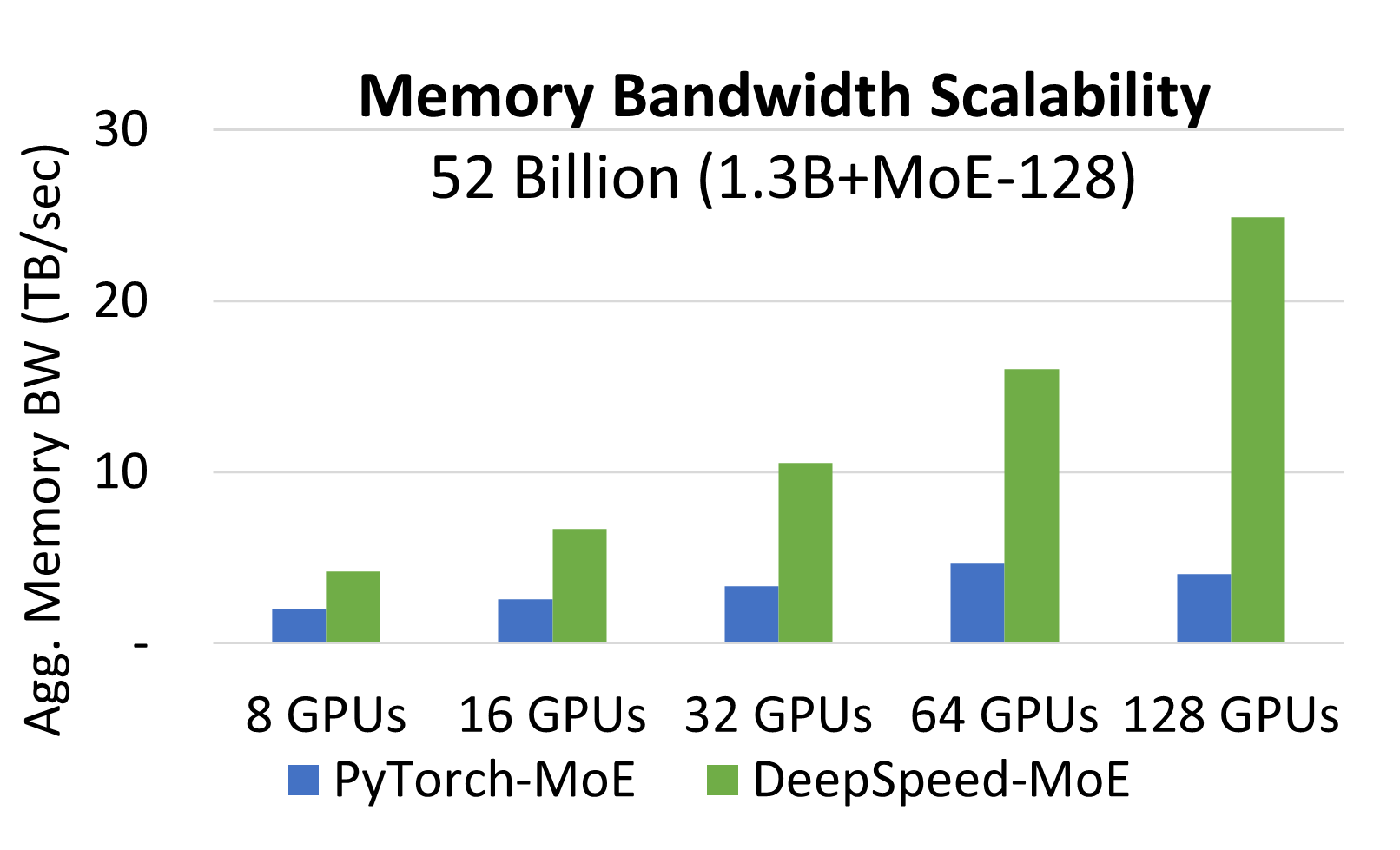}
\caption{Aggregate memory bandwidth scalability of \moe compared to baseline.}
\label{fig:perf-analysis-moe}
    \vskip -1em
\end{figure}

\begin{figure}[t]
\centering
\includegraphics[width=0.8\columnwidth]{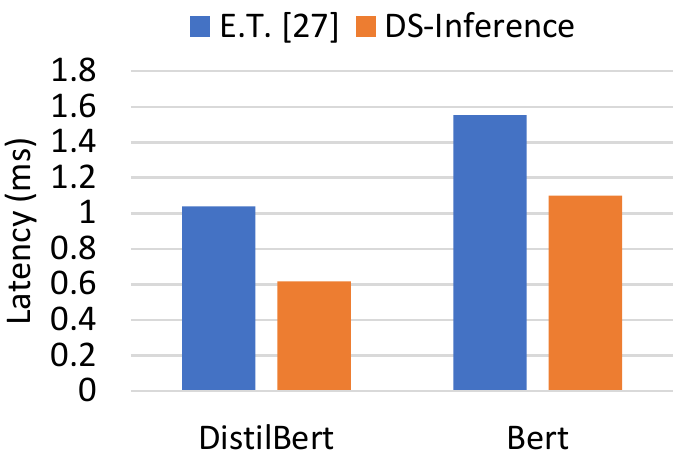}
\caption{Comparison with alternative Transformer kernels (E.T.~\cite{et-transformer}).}
\label{fig:comparison-et}
    \vskip -1em
\end{figure}

\subsection{Democratizing Larger Model Inference with \zinf}
\label{subsec:zero_inf_eval}
We evaluate three aspects of \zinf: 

\subsubsection{Model Scale} \zinf can inference a 530B parameter model on a single A6000 GPU, $25\times$ larger than the largest model that can be inferenced with a GPU-only solution (and $10\times$ larger compared to the CPU-only solution),  making it possible for data-scientists to test massive models on single GPU workstations without requiring massive GPU clusters or incurring huge cost (see Fig.~\ref{fig:zero_inf}(b)).

\subsubsection{Inference Throughput} \zinf achieves excellent inference throughput of up to $84$ TFLOPS, $54\%$ of theoretical peak (158.4 TFLOPS) for offline inference with very large batch sizes (see Fig.~\ref{fig:zero_inf}(b)). In fact, for models that fit in CPU memory, it offers over $25\times$ higher throughput than the CPU-only solution. Furthermore, even for models that fit in single GPU memory, it offers over $50\%$ better throughput than the GPU-only solution.
This is possible, because \zinf can support much larger batch sizes than a GPU-only solutions by offloading the parameters to CPU or NVMe and using GPU memory to store activations. The benefit of larger batch size is shown in Fig.~\ref{fig:zero_inf}(a).

\subsubsection{Scalability} When additional GPUs are available, \zinf can leverage them in parallel to achieve near perfect linear throughput (see Fig.~\ref{fig:zero_inf} (c)) by leveraging the aggregate PCIe bandwidth across GPUs as described in Sec.~\ref{subsubsec:zero_inf_efficiency}.

\subsection{Performance Breakdown and Analysis}
\label{s:perf-breakdown}

%

\subsubsection{Dense GPU kernel performance breakdown}Fig.~\ref{fig:perf-analysis}(a) shows that compared to PyTorch baseline, deep-fusion offers a significant reduction in latency by reducing kernel launch and data movement overheads, while our custom GeMM implementation offers further reduction for small batch sizes by increasing memory bandwidth utilization of GeMM.

\subsubsection{Throughput breakdown for massive model GPU-Inference} Fig.~\ref{fig:perf-analysis}(b) shows the impact of several optimizations in \OURS to the inference throughput, such as the dense optimized kernel, inference optimized scheduling, memory optimizations that lead to increased batch size, communication optimizations that reduce PCIe data movement overheads as described in Sec.~\ref{sec:many-gpu-dense}. 

\subsubsection{Prompt latency improvement with hybrid scheduling}
\begin{figure}[t]
\centering
\includegraphics[width=0.9\columnwidth]{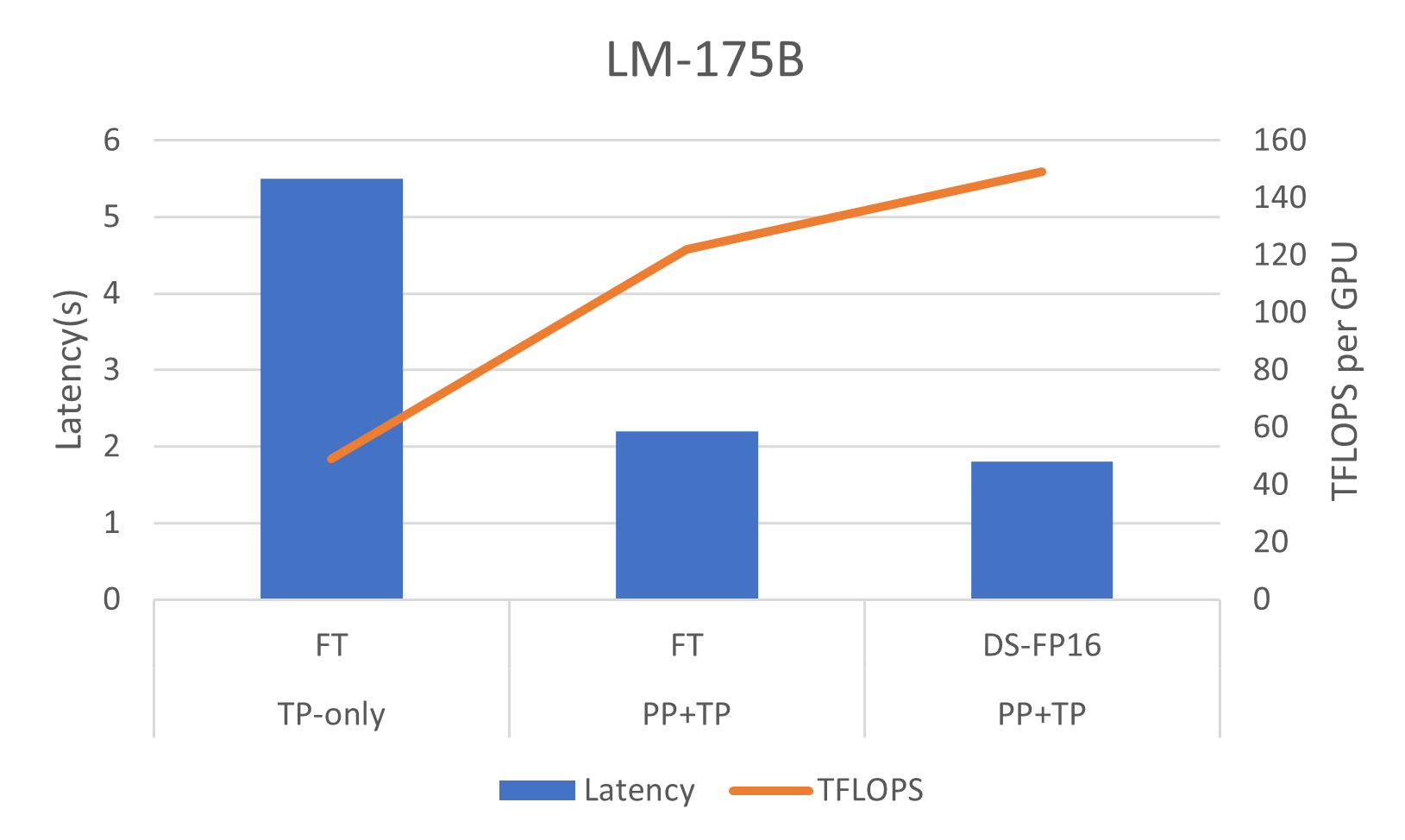}
\caption{Prompt processing latency and TFLOPS comparison between DeepSpeed with hybrid scheduling and  FasterTransformer.}
\label{fig:hybrid_perf}
    \vskip -1em
\end{figure}

Fig.\ref{fig:hybrid_perf} shows that DeepSpeed Inference with hybrid scheduling achieves 1.18x and 3.06x prompt processing speed-up over FasterTransformer for GPT-3 175B model with PP + MP configuration and MP-only configuration, respectively. This experiment was conducted on two nodes each with 8 A100 GPUs. We enable both pipeline and tensor parallelism. We set the batch size to 24, because the latency dramatically increases when the batch size is larger than 24. We suspect this is related to an issue in the AllReduce kernel in Pytorch. The results demonstrate hybrid scheduling has the potential to reduce prompt processing latency and we leave fixing the AllReduce issue as future work.


\subsubsection{Memory bandwidth scalability for sparse MoE models} Fig.~\ref{fig:perf-analysis-moe} shows that \OURS achieves much higher per GPU memory bandwidth than PyTorch baseline for a 52B MoE models on an $8\times$A100-GPU node while also demonstrating significantly better memory bandwidth scalability all the way to 128 GPUs that leads to the faster sparse model inference latency and higher throughput. \revision{This is the combined effect of MoE kernels and all-to-all optimizations presented in Section~\ref{sec:optimizing_moe_inference_latency}}. 

\subsubsection{Impact of pre-fetching on \zinf throughput} 
Fig.~\ref{fig:perf-analysis}(c) shows that prefetching (Sec.~\ref{subsubsec:zero_inf_efficiency}) improves throughput at small batch sizes while the benefit diminishing at larger batch sizes dues to higher arithmetic intensity to hide the CPU/NVMe to GPU communication overhead.

\subsubsection{Comparison with E.T.} \revision{We also compared with a state-of-the-art transformer kernel E.T.~\cite{et-transformer} for smaller scale DistilBERT and BERT encoder models on NVIDIA A100 GPUs for a batch size 1 and sequence length 128. Fig.~\ref{fig:comparison-et} shows that \OURS is 1.7x and 1.4x faster than E.T. on those two models. \OURS achieves lower latency because DeepFusion fuses more operators, leading to lower kernel invocation overhead and higher memory bandwidth utilization. In addition to being faster for small encoder models, we remark that the scope of our work is also much broader than E.T., where \OURS supports encoder, decoder, and sparsely gated MoE models at much larger scale.}

\section{Conclusion}
\label{sec:conclusion}

This paper presents \OURS, a system that enables efficient inference of transformer models at unprecedented scale, with respect to model size, the number of GPUs, and performance.
With innovations across the entire system stack, \OURS delivers speedy, efficient and economic inference as the model size grows, model architecture evolves, or the latency requirements become more stringent, supporting the increasing diversity of the transformer models and their application scenarios.  \OURS offers previously unattainable low latencies at unprecedented model scales, and make these gigantic models servable with unimaginably few resources. With such capabilities, we hope \OURS will not only facilitate the fast pace of innovation in transformer models but also further the state of using these models in production and research for everyone in need.

\newpage

\bibliographystyle{IEEEtran}
\bibliography{IEEEabrv,references}

\end{document}